\documentclass{article}

\usepackage{graphicx}
\usepackage{subfig}
\usepackage{booktabs} 

\usepackage{hyperref}

\usepackage[accepted]{icml2024}

\usepackage{amsmath}
\usepackage{amssymb}
\usepackage{mathtools}
\usepackage{amsthm}
\usepackage[ruled,vlined]{algorithm2e}
\usepackage{multirow}
\usepackage{xcolor,colortbl}
\usepackage{microtype}
\usepackage[american]{babel}
\usepackage{wrapfig}
\newcommand{\ie}{{\emph{i.e.}}}
\newcommand{\eg}{{\emph{e.g.}}}
\newcommand{\wrt}{{\emph{w.r.t.}}}
\usepackage{mdframed}
\newmdenv[
  topline=false,
  bottomline=false,
  rightline=false,
  linewidth=2pt,
  linecolor=black,
  leftmargin=0pt,
  rightmargin=0pt,
  innertopmargin=5pt,    
  innerbottommargin=5pt, 
  innerrightmargin=0pt,
  innerleftmargin=10pt, 
  skipabove=\medskipamount,
  skipbelow=2pt
]{findingbox}
\usepackage[capitalize,noabbrev]{cleveref}
\crefname{section}{Sec.}{Secs.}
\Crefname{section}{Section}{Sections}
\Crefname{table}{Table}{Tables}
\crefname{table}{Tab.}{Tabs.}
\crefname{equation}{Eq.}{Eqs.}
\crefname{algorithm}{Alg.}{Algs.}
\crefname{figure}{Fig.}{Figs.}
\crefname{appendix}{App.}{Apps.}
\theoremstyle{plain}
\newtheorem{theorem}{Theorem}[section]

\theoremstyle{definition}
\newtheorem{definition}[theorem]{Definition}

\theoremstyle{remark}

\SetCommentSty{myCommentStyle}

\usepackage[textsize=tiny]{todonotes}

\icmltitlerunning{Task Groupings Regularization: Data-Free Meta-Learning with Heterogeneous Pre-trained Models}

\begin{document}

\twocolumn[
\icmltitle{Task Groupings Regularization:\texorpdfstring{\\}{}Data-Free Meta-Learning with Heterogeneous Pre-trained Models}



\icmlsetsymbol{equal}{*}

\begin{icmlauthorlist}
\icmlauthor{Yongxian Wei}{yyy}
\icmlauthor{Zixuan Hu}{yyy,comp}
\icmlauthor{Li Shen}{sch}
\icmlauthor{Zhenyi Wang}{xxx}
\icmlauthor{Yu Li}{hk}
\icmlauthor{Chun Yuan}{yyy}
\icmlauthor{Dacheng Tao}{comp}
\end{icmlauthorlist}

\icmlaffiliation{yyy}{Tsinghua Shenzhen International Graduate School, Tsinghua University, China}
\icmlaffiliation{comp}{College of Computing \& Data Science, Nanyang Technological University, Singapore}
\icmlaffiliation{sch}{School of Cyber Science and Technology, Sun Yat-sen University, China}
\icmlaffiliation{xxx}{University of Maryland, College Park, USA}
\icmlaffiliation{hk}{Department of Computer Science and Engineering, The Chinese University of Hong Kong, China}
\icmlcorrespondingauthor{Li Shen}{mathshenli@gmail.com}
\icmlcorrespondingauthor{Zhenyi Wang}{zwang169@umd.edu}
\icmlcorrespondingauthor{Chun Yuan}{yuanc@sz.tsinghua.edu.cn}

\icmlkeywords{Machine Learning, ICML}

\vskip 0.3in
]



\printAffiliationsAndNotice{}  

\begin{abstract}
Data-Free Meta-Learning (DFML) aims to derive knowledge from a collection of pre-trained models without accessing their original data, enabling the rapid adaptation to new unseen tasks.
Current methods often overlook the heterogeneity among pre-trained models, which leads to performance degradation due to task conflicts.
In this paper, we empirically and theoretically identify and analyze the model heterogeneity in DFML. We find that model heterogeneity introduces a heterogeneity-homogeneity trade-off, where homogeneous models reduce task conflicts but also increase the overfitting risk. Balancing this trade-off is crucial for learning shared representations across tasks.
Based on our findings, we propose Task Groupings Regularization that benefits from model heterogeneity by grouping and aligning conflicting tasks.
Specifically, we embed pre-trained models into a task space to compute \emph{dissimilarity}, and group heterogeneous models together based on this measure. Then, we introduce implicit gradient regularization within each group to mitigate potential conflicts. By encouraging a gradient direction suitable for all tasks, the meta-model captures shared representations that generalize across tasks.
Comprehensive experiments showcase the superiority of our approach in multiple benchmarks, effectively tackling the model heterogeneity in challenging multi-domain and multi-architecture scenarios. Code is available \href{https://github.com/WalkerWorldPeace/TGR}{here}.
\end{abstract}

\section{Introduction}
Data-Free Meta-Learning (DFML)~\cite{kwon2020repurposing,wang2022meta,hu2023architecture,hu2023learning} aims to derive knowledge from a collection of pre-trained models without accessing their original data, enabling the rapid adaptation to new unseen tasks. Traditional meta-learning methods~\cite{finn2017model, wang2020bayesian} assume access to a collection of tasks with available training and testing data. However, such data is often not accessible in real-world scenarios due to privacy concerns, security risks, or usage rights restrictions~\cite{chen2019data,truong2021data,zheng2023learn}. For instance, numerous pre-trained models from diverse domains are released on platforms like GitHub or Hugging Face without accompanying training data. This situation underscores the value of DFML: collect some pre-trained models with weaker generalization abilities, which likely originate from diverse domains online, to train a meta-model with enhanced generalization capacity for new tasks.

Current DFML methods predominantly focus on data recovery from pre-trained models. For example, PURER~\cite{hu2023architecture} uses model inversion to optimize a learnable dataset for each pre-trained model and samples pseudo tasks for meta-learning. BiDf-MKD~\cite{hu2023learning} employs generators with multiple black-box APIs, generating support and query sets for meta-learning. Nonetheless, these methods often overlook the heterogeneity among pre-trained models, raised from differences in training data, model architecture, and optimization algorithm. Simultaneously learning from heterogeneous pre-trained models\footnote{In this paper, we consider pre-trained models as distinct tasks. It means that the data recovered from pre-trained models form individual tasks; and also that pre-trained models resemble diverse tasks in meta-learning, on which the meta-model is trained.} can sometimes lead to performance degradation due to task conflicts~\cite{standley2020tasks,shi2021gradient,ye2021multi}.

\begin{figure}[tb]
\centering
\includegraphics[width=\linewidth]{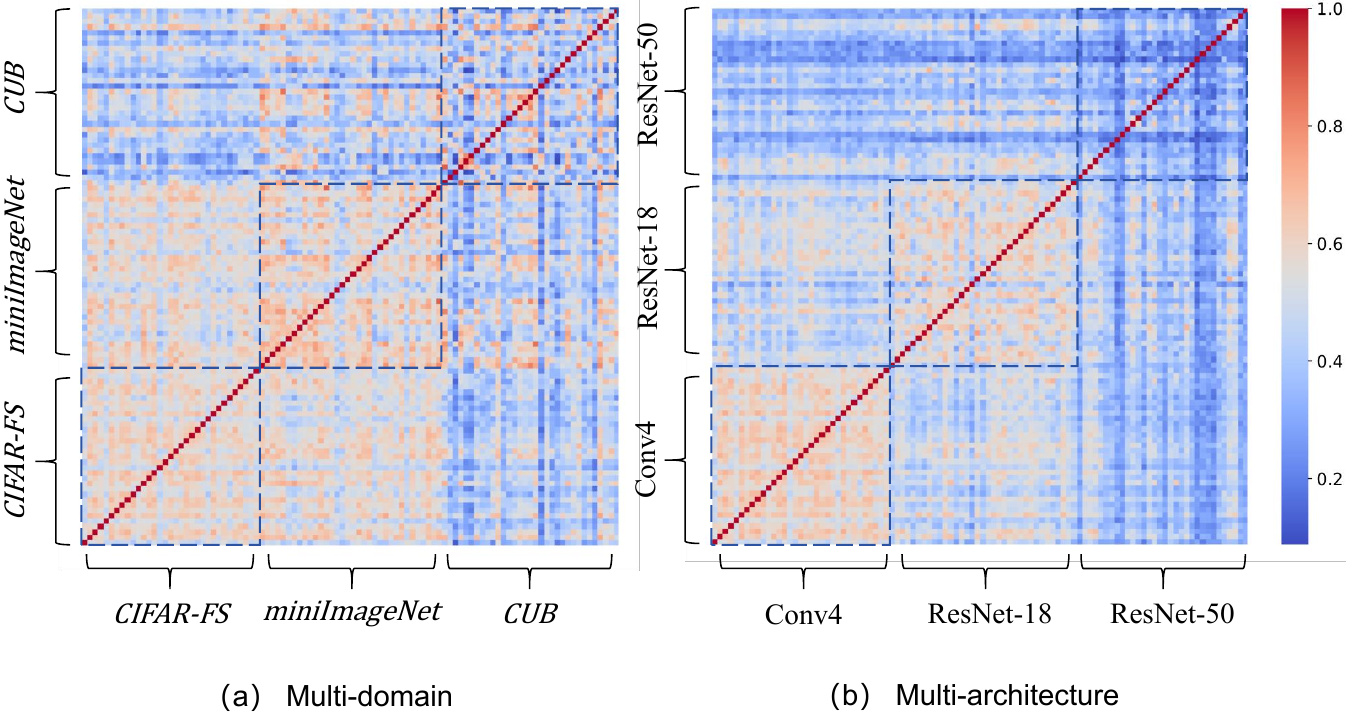}
\vspace{-1.5em}
\caption{\small Similarity heatmaps of pre-trained models measured by CKA. We compare (a) pre-trained models from different datasets, and (b) pre-trained models with different architectures. Coordinate axes indicate the corresponding model index, a total of 100 pre-trained models involved. Best viewed when zoomed in.}
\label{fig:CKA}
\vspace{-2em}
\end{figure}

In this paper, we first empirically demonstrate the existence and potential causes of model heterogeneity in DFML (see \cref{fig:CKA}), and analyze the practical effects of model heterogeneity on generalization (see \cref{sec:rethinking}). Experiments suggest that model heterogeneity is a double-edged sword and can be beneficial and desirable in certain cases (see \cref{fig:accuracy gain}). A certain degree of heterogeneity plays a role similar to regularization~\cite{kurin2022defense}, reducing the generalization error of the meta-model. Furthermore, we theoretically find that model heterogeneity introduces a heterogeneity-homogeneity trade-off. Conflicting tasks from heterogeneous models may compete for model capacity. Optimizing for task $\mathcal{T}_i$ could accidentally worsen the performance on task $\mathcal{T}_j$ and vice versa. In contrast, homogeneous models, while reducing task conflicts, also diminish data diversity due to their similar knowledge, thereby increasing the risk of overfitting. Balancing this trade-off is crucial for learning shared representations that generalize across tasks.

Based on our findings, we propose Task Groupings Regularization, a novel approach that benefits from model heterogeneity by balancing the heterogeneity-homogeneity trade-off. We divide the heterogeneous pre-trained models into task groups, each consisting of \emph{the most dissimilar} pre-trained models. Then, we introduce implicit gradient regularization within each group, to mitigate potential \emph{conflicts} of dissimilar tasks.
Specifically, we propose to embed pre-trained models into a task space to compute task dissimilarity. We first recover task-specific data from pre-trained models via model inversion. Subsequently, we utilize the Fisher Information Matrix (FIM)~\cite{amari1998natural} to extract task embeddings. Based on dissimilarity measures between these embeddings, we perform spectral clustering to divide models into multiple task groups. During the training phase of the meta-model, we periodically sample pre-trained models to inverse new tasks, facilitating the transfer of knowledge to the meta-model. Each time, we sample heterogeneous pre-trained models from the same task group. The tasks recovered from these heterogeneous models demonstrate conflicting optimization directions, failing to develop a shared representation. Therefore, we apply implicit gradient regularization among them, penalizing ``non-uniform'' regions to align conflicting tasks. By encouraging a gradient direction suitable for all tasks, the meta-model captures shared representations across different tasks from pre-trained models, enabling it to generalize effectively to unseen tasks.

Compared to existing DFML methods, our approach achieves improvements of 5.54\%, 6.70\% and 3.05\% on \emph{CIFAR-FS}, \emph{miniImageNet} and \emph{CUB}, respectively. Furthermore, we effectively tackle the model heterogeneity in challenging multi-domain and multi-architecture scenarios. 

In summary, our main contributions are three-fold:
\begin{itemize}
\item For the first time, we systematically identify and analyze the heterogeneity-homogeneity trade-off from both empirical and theoretical perspectives in DFML.
\item We propose Task Groupings Regularization, a novel approach that benefits from model heterogeneity. This approach aligns conflicting tasks, facilitating the learning of generalizable representations across tasks.
\item We conduct comprehensive experiments and discussions. Further on two challenging model heterogeneity scenarios, our empirical results demonstrate a significant improvement over previous methods.
\end{itemize}
\section{Related Work}
\label{sec:related}

\noindent
\textbf{Data-free meta-learning.}
Data-free learning~\cite{chen2019data,liu2021data,li2023deep} facilitates the learning process without access to the raw data. This approach is of significant value in practical scenarios where data availability is limited due to privacy, safety, or ethics. It has been greatly influenced by model inversion~\cite{fredrikson2015model}, which seeks to extract embedded data knowledge from pre-trained models.
Recently, Data-Free Meta-Learning (DFML)~\cite{wang2022meta,hu2023architecture,wei2024free,wei2024meta} has gained popularity, by leveraging multiple pre-trained models with weaker generalization abilities to learn a meta-model with superior generalization. DRO~\cite{wang2022meta} predicts meta initialization through a black-box neural network emphasizing the parameter space. PURER~\cite{hu2023architecture} progressively synthesizes a series of pseudo-tasks using episode curriculum inversion. BiDf-MKD~\cite{hu2023learning} learns the meta-model by transferring knowledge from black-box APIs via zero-order gradient estimation. However, they overlook the heterogeneity inherent in the collected pre-trained models.

\noindent
\textbf{Task groupings.}
It is commonly believed that related tasks with similar underlying structures can benefit from training together. In multi-task learning~\cite{caruana1993multitask}, deciding which tasks to train together has traditionally involved an extensive search~\cite{standley2020tasks} or subjective human judgment. Recently, TAG~\cite{fifty2021efficiently} determines task groupings by jointly training all tasks and assessing the impact of one task’s gradient on another task’s loss. To mitigate Negative Transfer~\cite{jiang2023forkmerge}, \citet{zamir2018taskonomy} suggest assigning unrelated tasks to different groups to minimize task interference. However, after establishing task groupings, most methods simply create a unified gradient by linearly combining gradients of tasks. Different from previous work, we propose grouping dissimilar tasks together and then applying regularization to align their gradients. Moreover, our task grouping is a pre-processing step, which does not increase the training overhead.

\noindent
\textbf{Gradient alignment.}
Gradient alignment~\cite{shi2021gradient, wang2022learning} highlights the importance of aligning or maintaining consistency in gradients during training. Empirical studies have demonstrated its effectiveness in various domains like meta-learning~\cite{eshratifar2018gradient}, federated learning~\cite{dandi2022implicit}, and multi-task learning~\cite{guangyuan2022recon}. In our work, we design an implicit gradient regularization to align tasks from heterogeneous models, with low computation time and memory overheads.

\section{Revisit DFML}

In this section, we first describe the basic definition of DFML and then revisit the model heterogeneity challenges.

\subsection{Problem Setup of DFML}
We are given only a collection of pre-trained models $\mathcal{M}_{pool}=\{M_i\}_{i=1}^{n}$ without accessing their training data. DFML utilizes synthetic data $\boldsymbol{\hat{X}}$ recovered from $\mathcal{M}_{pool}$ to train a meta-model $M_{meta}(\cdot;\boldsymbol{\theta})$, which can be rapidly transferred to unseen tasks. During meta-testing, we sample ``$N$-way $K$-shot'' tasks. For such a task, there are $N$ classes and each class only has $K$ labeled samples named the support set, and $U$ unlabelled samples per class called the query set. We use the few-shot support set to adapt the meta-model $M_{meta}(\cdot;\boldsymbol{\theta})$ to each specific task. The query set is then used to make predictions and assess the accuracy.
 
Next, we define the Accuracy Gain to measure the impact of joint training in DFML, using two pre-trained models.
\begin{definition}
\label{def:accuracy gain}
\textbf{(Accuracy Gain).} Denote the meta-model trained from a single pre-trained model as $\boldsymbol{\theta}(M_\mathrm{bas})$, and the meta-model trained from an auxiliary pre-trained model alongside the basic one as $\boldsymbol{\theta}(M_\mathrm{bas},M_\mathrm{aux})$. Let $\mathcal{P}$ be the meta-testing accuracy as a performance measure. Then, the Accuracy Gain of joint training can be evaluated by:
\begin{equation}\label{eq:accuracy gain}
AG = \mathcal{P} (\boldsymbol{\theta}(M_\mathrm{bas},M_\mathrm{aux})) - \mathcal{P} (\boldsymbol{\theta}(M_\mathrm{bas}) ) .
\end{equation}
\end{definition}

\subsection{Rethinking Model Heterogeneity in DFML}

\textbf{Model heterogeneity analysis.}\label{sec:rethinking}
The heterogeneity in collected pre-trained models is inevitable, arising from differences in training data, model architecture, and optimization algorithm, particularly concerning the degree of model convergence. Models pre-trained on varied data distributions can be distinguished based on their specific domains. Even within the same dataset, models trained by different classes exhibit distinct sub-domain characteristics, \eg, statistical properties like pixel intensity or texture variation~\cite{chen2021variational,song2020robust}. Moreover, the diversity in model architectures also contributes to model heterogeneity, influencing how data is represented~\cite{hermann2020shapes}. This in turn affects the patterns of the recovered data, leading to a gap in their distributions.

To demonstrate the heterogeneity among pre-trained models, we use Centered Kernel Alignment (CKA)~\cite{kornblith2019similarity,hao2023one} to measure the similarity of features extracted by various pre-trained models. CKA is well-suited for comparisons across different architectures since it can accommodate inputs of varying dimensions. The similarity heatmaps, presented in \cref{fig:CKA}, distinctly showcase the substantial differences in feature distribution among models. Heterogeneity arises from different datasets and model architectures. This is particularly obvious in the fine-grained \emph{CUB} dataset and the large-scale ResNet-50 architecture, showing increased internal heterogeneity.

\begin{figure}[t]
\begin{minipage}[t]{\linewidth}
        \vspace{-1em}
        \centering
        \subfloat[Overlapping classes]{\includegraphics[width=0.49\columnwidth]{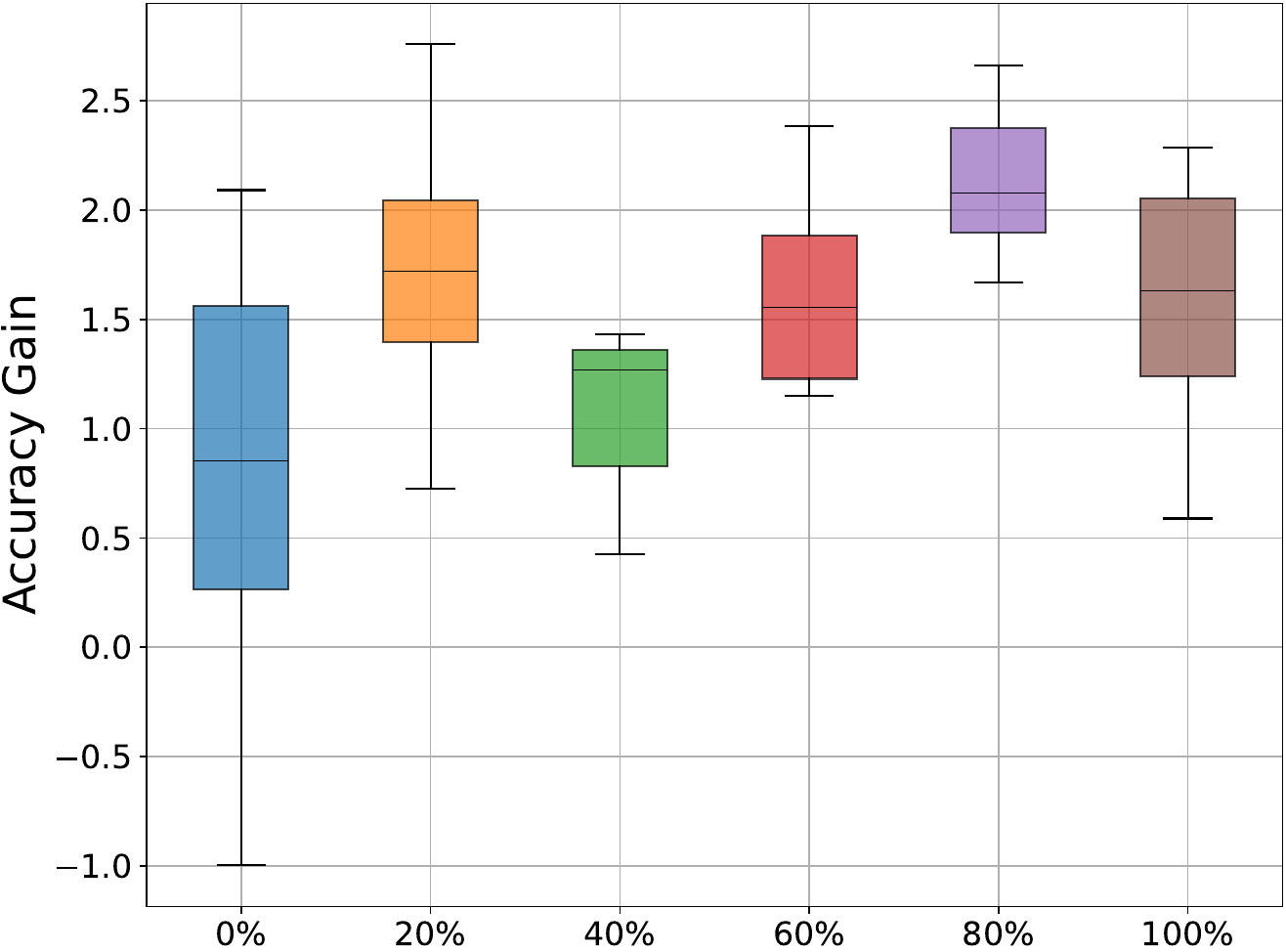} }
        \subfloat[Model architecture]{\includegraphics[width=0.49\columnwidth]{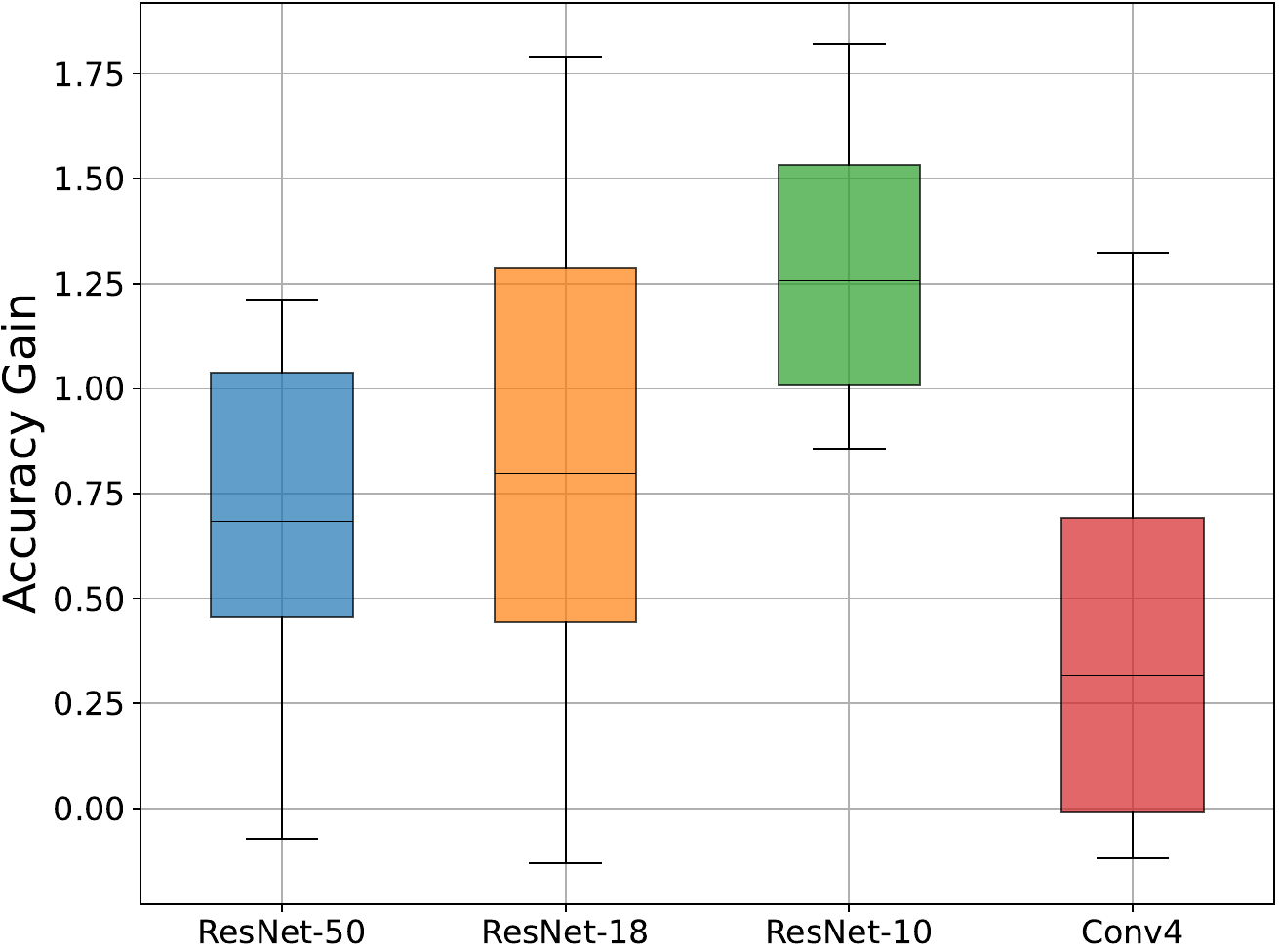} }
        \vspace{-0.5em}
         \caption{\small Relationships between the model heterogeneity and Accuracy Gain. We select a pre-trained Conv4 as the basic model, combining it with an auxiliary pre-trained model to assess the AG.}
	\label{fig:accuracy gain}
\end{minipage}
\vspace{-0.5em}
\end{figure}

\textbf{Empirical observation.}
To analyze effects of model heterogeneity based on the defined Accuracy Gain, we evaluate model heterogeneity using two metrics: the ratio of overlapping training classes and the architecture of pre-trained models. This is based on the reasonable assumption that a lower overlap in classes indicates differing task distributions, and the representation ways of different model architectures vary. We select a pre-trained Conv4 as the basic model $M_\mathrm{bas}$, combining it with various auxiliary models $M_\mathrm{aux}$ to assess the AG. For a fair comparison, each measure begins with the same model parameters. The process of training data recovery follows \cref{eq:Loss_g}, which we will discuss later. Detailed designs of our experiment are provided in \cref{app:rethink}.

\cref{fig:accuracy gain} demonstrates that combining homogeneous models does not yield significant AG, \eg, 100\% overlapping classes in (a), and the same Conv4 architecture in (b). Conversely, choosing heterogeneous models results in higher AG. This is attributed to the fact that while conflicting gradients from heterogeneous models will increase training loss and slow down convergence speed, they may also play a role similar to regularization~\cite{kurin2022defense}, reducing the overfitting risk, thereby reducing the generalization error of meta-model~\cite{jiang2023forkmerge}. However, this can also result in significant performance degradation when tasks compete for model capacity or fail to develop a shared representation that generalizes across tasks, \eg, 0\% overlapping classes in (a), and the ResNet-50 architecture in (b).

\begin{figure*}[t]
  \centering
    \includegraphics[width=0.95\linewidth]{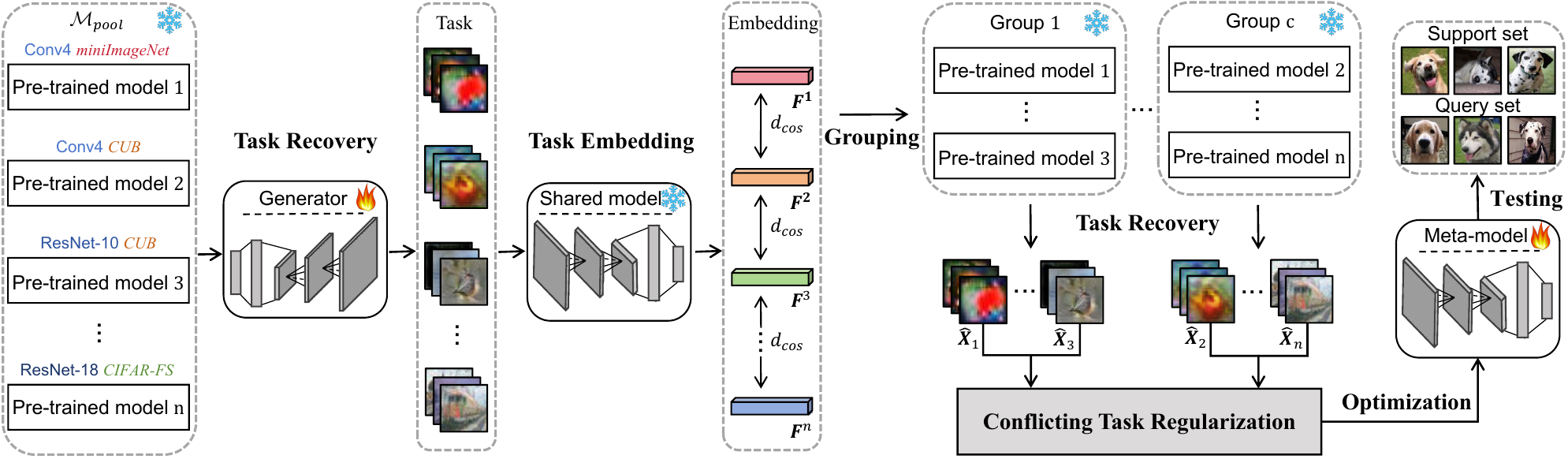}
    \vspace{-0.5em}
   \caption{DFML training pipeline. We embed pre-trained models $\mathcal{M}_{pool}$ into a task space to compute dissimilarity, and divide heterogeneous models into task groups. Then, conflicting task regularization is introduced within each group to train the meta-model for new tasks.}
   \label{fig:method}
\end{figure*}

\textbf{Theoretical understanding.}
Inspired by the empirical observation, we will argue that there is a trade-off between minimizing the expected error over the merged distribution via the empirical error of joint learning.
\begin{theorem}\label{theorem0}
Assume $M_{meta}(\cdot;\boldsymbol{\theta})$ is probably approximately correct (PAC), \ie, there exists $\zeta(N,\delta)\geq0$ monotonically decreasing with $N$, and the loss function $\ell(\cdot)$ is $K$-Lipschitz continuous. Then, with probability at least $1-2 \delta$ the following bounds hold:
\begin{align*}
&|E(\boldsymbol{\theta}(M_\mathrm{bas}),\boldsymbol{\theta}(M_\mathrm{aux}))-
\hat{E}(\boldsymbol{\theta}(M_\mathrm{bas},M_\mathrm{aux}))| \le \zeta(N,\delta)\\
& +\!K\!\!\sum_{t=\mathrm{bas}}^{\mathrm{aux}}\!\!\mathbb{E}_{\mathcal{P}_t}[|M_{meta}(\boldsymbol{\theta}(M_t))\!-\!M_{meta}(\boldsymbol{\theta}(M_\mathrm{bas},M_\mathrm{aux}))|]
\end{align*}
where $E=\sum_{t=\mathrm{bas}}^{\mathrm{aux}}\mathbb{E}_{\mathcal{P}_t}[\ell\left(M_{meta}(x,\boldsymbol{\theta}(M_t))\right)]$ is the expected error over the merged distribution and $\hat{E}=\frac{1}{N}\sum_{i=1}^{N}\ell\left(M_{meta}(x_i,\boldsymbol{\theta}(M_\mathrm{bas},M_\mathrm{aux}))\right)$ is the empirical error over the $N$ training samples.
\end{theorem}
The bound consists of two terms: a homogeneity term $\zeta(N,\delta)$ dependent on the number of samples $N$; and one term $M_{meta}(\boldsymbol{\theta}(M_t))-M_{meta}(\boldsymbol{\theta}(M_\mathrm{bas},M_\mathrm{aux}))$ based on the distance between meta-models trained from different models. This term can be seen as a heterogeneity term, arising from the discrepancy between $M_\mathrm{bas}$ and $M_\mathrm{aux}$. When the distributions of $M_\mathrm{bas}$ and $M_\mathrm{aux}$ are close, a significant portion of the data recovered from them overlap~\cite{yin2020dreaming}, \ie, a reduction in the number of distinct data sampled from the merged distribution $\mathcal{P}_\mathrm{bas}+\mathcal{P}_\mathrm{aux}$, thereby increasing the homogeneity term. On the other hand, when model heterogeneity is substantial, the discrepancy between $\boldsymbol{\theta}(M_\mathrm{bas})$ and $\boldsymbol{\theta}(M_\mathrm{aux})$ becomes significant~\cite{cervino2023multi}, making the heterogeneity term larger. Thus, \cref{theorem0} highlights a heterogeneity-homogeneity trade-off that necessitates careful handling.
\begin{findingbox}
\textbf{Finding.} \emph{Model heterogeneity presents a heterogeneity-homogeneity trade-off, and balancing this trade-off is crucial for the meta-model's generalization across tasks.}
\end{findingbox}

Consequently, it is a natural consideration to identify task groups, \ie, decide which models should be trained together to balance the trade-off. The decision of task groupings is complex and not thoroughly explored, often relying on the judgment of human experts~\cite{zhang2021survey}.

\section{Methodology}
Guided by our findings, we propose Task Groupings Regularization (see \cref{fig:method}) to learn from a collection of pre-trained models $\mathcal{M}_{pool}$. Initially, we recover specific tasks from these models via a generator, and use them to compute the FIM as task embeddings. Based on dissimilarity of these embeddings, we group heterogeneous models together. Then, implicit gradient regularization is applied within each group to mitigate potential conflicts among tasks.

Below, we describe the two parts in detail. In \cref{sec:group}, we divide $\mathcal{M}_{pool}$ into $c$ task groups $\mathcal{G}_i \subset \mathcal{M}_{pool}$, each consisting of the most dissimilar pre-trained models. Then in \cref{sec:conflicting task}, we introduce conflicting task regularization to each group for optimizing the meta-model.

\subsection{Heterogeneous Pre-trained Models Grouping}\label{sec:group}
Our understanding of the heterogeneity in pre-trained models is limited since the original data is unknowable. Moreover, due to the exponential increase in the number of group combinations $2^{n}-1$ with the number of pre-trained models $n$, it becomes both challenging and computationally demanding to group these models directly. 
To address these, we propose to embed pre-trained models into a task space to compute task dissimilarity. By performing spectral clustering based on the dissimilarity measure, we successfully group heterogeneous models without the need for training.

\textbf{Task recovery via model inversion.}
Given a pre-trained model $M$, we optimize a generator $G(\cdot;\boldsymbol{\theta}_G)$ alongside a corresponding latent code $\boldsymbol{Z}$ to recover training data $\boldsymbol{\hat{X}}$. These synthetic data $\boldsymbol{\hat{X}}$ mimic similar responses from the pre-trained model as the real data $\boldsymbol{X}$. For recovering the task from pre-trained model $M$, we employ the inversion loss to update $\boldsymbol{\theta}_G$ and $\boldsymbol{Z}$, which comprises a classification loss $\mathcal{L}_{CE}$ and a feature regularization loss $\mathcal{L}_{BN}$:
\begin{equation}\label{eq:Loss_g}
\min_{\boldsymbol{Z},\boldsymbol{\theta}_{G}}\mathcal{L}_{G}(\boldsymbol{Z},\boldsymbol{Y};\boldsymbol{\theta}_{G})= \mathcal{L}_{CE}+\mathcal{L}_{BN},
\end{equation}
where $\mathcal{L}_{CE}$ is a cross-entropy loss that conditions the input data on pre-defined task labels $\boldsymbol{Y}$. This enforces the synthetic data $\boldsymbol{\hat{X}}=G(\boldsymbol{Z};\boldsymbol{\theta}_G)$ to yield strong predictions~\cite{chen2019data}, akin to a one-hot vector:
\begin{equation}\label{eq:Loss_ce}
\mathcal{L}_{CE}(\boldsymbol{\hat{X}},\boldsymbol{Y};\boldsymbol{\theta}_{G})= {CE}(M(\boldsymbol{\hat{X}}),\boldsymbol{Y}).
\end{equation}
$\mathcal{L}_{BN}$ ensures the alignment of feature-map statistics of the synthetic data with those of the real training data~\cite{yin2020dreaming}, as preserved in the Batch Normalization (BN) layers of the pre-trained models:
\begin{equation}\label{eq:Loss_bn}
\!\!
\mathcal{L}_{BN}(\boldsymbol{\hat{X}};\boldsymbol{\theta}_{G})\!=\!\sum_{l}\|\mu_{l}(\boldsymbol{\hat{X}})\!-\!\mu_{l}^{\mathrm{BN}}\|\!+\!\|\sigma_{l}^{2}(\boldsymbol{\hat{X}})\!-\!\sigma_{l}^{\mathrm{BN}}\|,\!\!
\end{equation}
where $\mu_{l}(\cdot)$ and $\sigma_{l}^{2}(\cdot)$ denote their mean and variance at the $l^\mathrm{th}$ convolutional layer. $\mu_{l}^{\mathrm{BN}}$ and $\sigma_{l}^{\mathrm{BN}}$ refer to the mean and variance of prior statistics stored in the $l^\mathrm{th}$ BN layer.

We recover $\boldsymbol{\hat{X}}$ that minimize \cref{eq:Loss_g} by optimizing $(\boldsymbol{Z},\boldsymbol{\theta}_G)$. In the absence of real tasks, we generate a certain amount of data (600 images per class in our experiment) to form a pseudo task that represents each pre-trained model.

\textbf{Task grouping via dissimilarity measures.}
Fisher Information Matrix (FIM) serves as an indicator of a model's sensitivity to minor perturbations and the curvature of its loss surface. Furthermore, the FIM can be leveraged as the task embedding, capturing task domain and task difficulty~\cite{achille2019task2vec}. Formally, the empirical FIM is defined as the covariance of the gradients of the log-likelihood with respect to model parameters ${\boldsymbol{\varphi}}$:
\begin{equation}\label{FIM}
\!\!\boldsymbol{F}_{\boldsymbol{\varphi}}^i
\!=\!\frac{1}{N}\sum_{j=1}^{N}\!\left[\nabla_{\boldsymbol{\varphi}}\log P_{\boldsymbol{\varphi}}(y_j|x_j)\nabla_{\boldsymbol{\varphi}}\log P_{\boldsymbol{\varphi}}(y_j|x_j)^\mathrm{T}\right],\!
\end{equation}
where $P_{\boldsymbol{\varphi}}(y|x)$ is the prediction function parametrized by ${\boldsymbol{\varphi}}$. We utilize the FIM to compute the embedding of task $\{(x_j,y_j)\}_{j=1}^N$ recovered from pre-trained model $M_i$.

The FIM offers a fixed-dimensional representation of a given task, unaffected by factors like the number of classes. We can select a shared pre-trained model and freeze its weights $\boldsymbol{\varphi}$ to compute the task embedding $\boldsymbol{F}$. Given the large dimensionality of $\boldsymbol{F}_{\boldsymbol{\varphi}}$, we approximate by considering only its diagonal entries, following the method of \citet{wu2023pi}. Thus, the task embedding $\boldsymbol{F}$ is computed as $\boldsymbol{F}=\operatorname{diag}(\boldsymbol{F}_{\boldsymbol{\varphi}})=[\boldsymbol{F}_{\boldsymbol{\varphi}_1},\boldsymbol{F}_{\boldsymbol{\varphi}_2},\cdots,\boldsymbol{F}_{\boldsymbol{\varphi}_n}]$, where $\boldsymbol{\varphi}_i$ represents the $i^\mathrm{th}$ parameter of the shared model.

We then calculate the cosine similarity between the embeddings of each pair of tasks to obtain a similarity matrix (see \cref{app:similar}). By subtracting this matrix from an all-ones matrix $\boldsymbol{J}$, we derive the task dissimilarity matrix $\boldsymbol{W}$:
\begin{equation}\label{eq:dissimilarity}
\boldsymbol{W}=\boldsymbol{J} - \left[ d_{\cos}\Big(\frac{\boldsymbol{F}^i}{\boldsymbol{F}^i+\boldsymbol{F}^j},\frac{\boldsymbol{F}^j}{\boldsymbol{F}^i+\boldsymbol{F}^j}\Big)\right]_{1 \leq i, j \leq n}.
\end{equation}
The measures reflect the relations of tasks, correlating positively with natural distances in the task space. This allows us to identify task groupings that maximize total dissimilarity measures. While this problem is NP-hard (reduction from Graph Partitioning), it can be efficiently addressed using a clustering algorithm like spectral clustering~\cite{ng2001spectral}. With the pre-computed $\boldsymbol{W}$, the objective function of spectral clustering is defined by:
\begin{equation}\label{eq:clustering}
\mathop{\arg\min_{\boldsymbol{H}}}~Tr(\boldsymbol{H}^\top\boldsymbol{L}\boldsymbol{H}),\,
\mathrm{s.t.}~\boldsymbol{H}^\top\boldsymbol{H}=\boldsymbol{I},
\end{equation}
where $\boldsymbol{L}$ represents the Laplacian matrix defined by $\boldsymbol{L} = \boldsymbol{D}-\boldsymbol{W}$, $\boldsymbol{D}$ is a diagonal matrix with $\boldsymbol{D}_{ii}=\sum_j\boldsymbol{W}_{ij}$, and $Tr(\cdot)$ denotes the
trace of a matrix. The optimal data representation $\boldsymbol{H}$ to \cref{eq:clustering} comprises $c$ eigenvectors corresponding to the $c$ smallest eigenvalues of $\boldsymbol{L}$. With the optimal $\boldsymbol{H}$, the clustering assignment is obtained by performing $k$-means on it, dividing $\mathcal{M}_{pool}$ into $c$ groups. The spectral clustering algorithm integrates the global relationships between tasks (\ie, dissimilarity matrix $\boldsymbol{W}$), and performs clustering in the embedding space of tasks that are linearly inseparable~\cite{huang2019multi}. Dissimilar pre-trained models are grouped together, which diversifies recovered samples, thereby reducing the homogeneity term.

\subsection{Conflicting Task Regularization}\label{sec:conflicting task}
During the training phase of the meta-model, we periodically sample pre-trained models to inverse new tasks, facilitating the transfer of knowledge to the meta-model. Each time, we sample heterogeneous pre-trained models from the same group. Tasks from heterogeneous models demonstrate conflicting optimization directions, as evidenced by their gradients inner product $<0$. For optimal generalization performance, it's crucial to not just converge to a solution that minimizes the mean loss, but also to steer towards regions where gradient discrepancy is minimized.

\textbf{Meta-model optimization.}
For pre-trained model $M_i$, we recover the pseudo task $\hat{\boldsymbol{X}_{i}}$ as outlined in \cref{eq:Loss_g}. This pseudo task is then utilized for knowledge distillation, allowing for the transfer of task-specific knowledge from the pre-trained model (the teacher) to the meta-model (the student). The optimization of the meta-model focuses on minimizing the discrepancies in predictions between the teacher $M_i$ and student $M_{meta}$. In addition to using soft-label predictions, we also implement hard-label supervision with pre-defined task labels $\boldsymbol{Y}$ to compute the loss $\mathcal{L}_i(\boldsymbol{\theta})$:
\begin{equation}\label{eq:loss_kt}
\begin{aligned}
\mathcal{L}_i(\boldsymbol{\theta})
= &KL(M_{i}(\hat{\boldsymbol{X}_{i}}),M_{meta}(\hat{\boldsymbol{X}_{i}};\boldsymbol{\theta}))\\
&+CE(M_{meta}(\hat{\boldsymbol{X}_{i}};\boldsymbol{\theta}),\boldsymbol{Y}).
\end{aligned}
\end{equation}
Once the newly recovered task $\hat{\boldsymbol{X}_{i}}$ has undergone knowledge transfer, it is stored in a memory bank $\mathcal{B}$. We introduce cross-task replay to broaden the pseudo task distribution. Specifically, we sample from $\mathcal{B}$ to obtain tasks that are interpolated across different tasks, \ie, a random combination of classes. For each interpolated task $\hat{\mathcal{T}}$, we divide it into a support set and a query set. Then, we utilize MAML~\cite{finn2017model}, as described in \cref{eq:maml} in the Appendix, to optimize the meta-model. Broadening the range of training tasks is crucial for effective meta-learning.

\begin{algorithm}[tb]
\small
\DontPrintSemicolon
\SetKwInOut{Input}{Input}\SetKwInOut{Output}{Output}\SetKwInOut{Require}{Require}
\textbf{Input: }{A collection of pre-trained models $\mathcal{M}_{pool}$; a shared model~$\boldsymbol{\varphi}$; a generator $G(\cdot;\boldsymbol{\theta}_G)$; the memory bank $\mathcal{B}$.
}\\
\textbf{Output: }{A meta-model $M_{meta}(\cdot;\boldsymbol{\theta})$ for new tasks.}\\
\tcp{Pre-trained Models Grouping}
Task recovery via $G(\cdot;\boldsymbol{\theta}_G)$ \wrt \, \cref{eq:Loss_g}\\
Task embedding via $\boldsymbol{\varphi}$ \wrt \, \cref{FIM}\\
Grouping via dissimilarity measures \wrt \, \cref{eq:dissimilarity,eq:clustering}\\
\For{ {\rm epoch} $ \leftarrow 1$ \KwTo $T$}{
\tcp{Conflicting Task Regularization}
Sample a task group $\mathcal{G}_i$ from $\mathcal{M}_{pool}$\\
Sample $m$ pre-trained models from $\mathcal{G}_i$\\
Recover new tasks $\{\boldsymbol{\hat{X}}_{i}\}_{i=0}^{m-1}$ via $G(\cdot;\boldsymbol{\theta}_G)$\\
Compute the loss $\mathcal{L}_i(\boldsymbol{\theta})$ for each task \wrt \, \cref{eq:loss_kt}\\
Optimize $M_{meta}(\cdot;\boldsymbol{\theta})$ with $\frac{1}{m}\sum_{i=0}^{m-1}\nabla {\mathcal{L}_i}(\boldsymbol{\theta}-\boldsymbol{v}_i(\boldsymbol{\theta}))$\\
\tcp{Cross-Task Replay}
Put into memory bank $\mathcal{B} \leftarrow \mathcal{B} + \{\boldsymbol{\hat{X}}_{i}\}_{i=0}^{m-1}$\\
Construct interpolated tasks $\hat{\mathcal{T}}$ from $\mathcal{B}$\\
Optimize the meta-model $M_{meta}$ with $\hat{\mathcal{T}}$ \\
}
\caption{Task Groupings Regularization}
\label{alg:pipeline}
\end{algorithm}

\textbf{Implicit gradient regularization among tasks.}
Regularization broadly refers to methods involving penalty functions that enhance model generalization~\cite{zhao2022penalizing}. \citet{smith2020origin} discovered that the mean SGD iterate remains close to the path of gradient flow on a loss that includes an implicit regularizer $\sum_{i=0}^{m-1}||\nabla{\mathcal{L}_i}(\boldsymbol{\theta})-\nabla \bar {\mathcal{L}}(\boldsymbol{\theta})||^2$, where $\nabla \bar {\mathcal{L}}(\boldsymbol{\theta})=\frac{1}{m}\sum_{i=0}^{m-1}\nabla{\mathcal{L}_i}(\boldsymbol{\theta})$ is the average gradient over minibatches. This regularizer penalizes ``non-uniform'' regions characterized by large norms of errors in the minibatch gradients. However, this assumption only holds true when the learning rate is small and finite.

It becomes intuitive to consider the explicit implementation of this regularizer, which measures the variance of gradients across conflicting tasks within a group. Minimizing this variance facilitates the alignment of these tasks. Nevertheless, optimizing it involves calculating the Hessian matrix. For DNNs, directly computing such a large-dimensional Hessian matrix entails prohibitive time and memory overheads. Therefore, an appropriate approximation method should be implemented in this calculation. \citet{dandi2022implicit} point out that gradient alignment in sequential updates for SGD originates from evaluating the gradient of a minibatch $i$ after an additional displacement in the direction of $-(\nabla \bar {\mathcal{L}}(\boldsymbol{\theta})-\nabla{\mathcal{L}_i}(\boldsymbol{\theta}))$. Consequently, we can mimic this gradient alignment characteristic by using gradients for each task $i$ that are computed following an initial displacement given by $\boldsymbol{v}_i(\boldsymbol{\theta})=\beta\left(\nabla \bar {\mathcal{L}}(\boldsymbol{\theta})-\nabla {\mathcal{L}_i}(\boldsymbol{\theta})\right)$.

\begin{theorem}\label{theorem1}
If ${\mathcal{L}_i}(\boldsymbol{\theta})$ has Lipschitz Hessian, then the update gradient $\boldsymbol{g}_{IGR}=\frac{1}{m}\sum_{i=0}^{m-1}\nabla {\mathcal{L}_i}(\boldsymbol{\theta}-\boldsymbol{v}_i(\boldsymbol{\theta}))$, calculated following an initial displacement applied to the parameters $\boldsymbol{\theta}$, can be expressed as follows:
\begin{align*}
\!\boldsymbol{g}_{IGR}\!=\!\nabla \bar {\mathcal{L}}(\boldsymbol{\theta}) \!+\! \underbrace{\frac{\beta}{2m}\nabla(\sum_{i=0}^{m-1}\|\nabla \mathcal{L}_i(\boldsymbol{\theta})\!-\!\nabla \bar {\mathcal{L}}(\boldsymbol{\theta})\|^2)}_{Gradient\,Regularization}\!+\mathcal{O}(\beta^2),
\end{align*}
\ie, implicitly minimizes the trace of the covariance matrix for the conflicting task gradients, along with the Empirical Risk Minimization (ERM) gradient on the mean loss $\bar {\mathcal{L}}(\boldsymbol{\theta})$.
\end{theorem}

The above theorem implicitly penalizes ``non-uniform'' regions to align conflicting tasks, minimizing the gradient discrepancy (effects of implicit gradient regularization are validated in \cref{fig:gdm}). It guides the meta-model towards a unified direction for learning task-shared representations. Thus, we mitigate the discrepancy of $\boldsymbol{\theta}$ trained from different pre-trained models, reducing the heterogeneity term.

To conclude, we concisely outline the pipeline of the proposed framework in \cref{alg:pipeline}.

\section{Experiments}
In this section, we empirically verify the effectiveness of our proposed approach via extensive experiments.

\begin{table*}[tb]
\footnotesize
\centering
\renewcommand\arraystretch{1.1}
\setlength{\tabcolsep}{4.85pt}
\caption{\small Compare to existing baselines in DFML. We report on two categories of DFML algorithms: the non-inversion methods and the inversion-based methods. All results are the average accuracy with 95\% confidence intervals. The best results are shown in bold.}
\resizebox{0.82\linewidth}{!}{
\begin{tabular}{lcccccc}
			\addlinespace
			\toprule
			\specialrule{0em}{1pt}{1pt}
			\multirow{2}{*}{ \bf Method} &\multicolumn{2}{c}{\emph {CIFAR-FS}} & \multicolumn{2}{c}{\emph {miniImageNet}} & \multicolumn{2}{c}{\emph {CUB}}\\ 
			\cmidrule(l){2-3} \cmidrule(l){4-5} \cmidrule(l){6-7}
			\specialrule{0em}{1pt}{1pt}
			&{$5$-way $1$-shot} & {$5$-way $5$-shot} &{$5$-way $1$-shot} &{$5$-way $5$-shot} &{$5$-way $1$-shot} &{$5$-way $5$-shot}\\
			\midrule
			\specialrule{0em}{1pt}{1pt}
			Finetune  & {28.59} {\scriptsize $\pm$ 0.56} & {34.77} {\scriptsize $\pm$ 0.62} &{25.06} {\scriptsize $\pm$ 0.50}& {28.10} {\scriptsize $\pm$ 0.52}  &{26.26} {\scriptsize $\pm$ 0.48}& {29.89} {\scriptsize $\pm$ 0.55}\\
			\specialrule{0em}{1pt}{1pt}
           Best-Model  & {21.68} {\scriptsize $\pm$ 0.66} & {25.05} {\scriptsize $\pm$ 0.67} & {22.86} {\scriptsize $\pm$ 0.61}& {26.26} {\scriptsize $\pm$ 0.63} & {24.16} {\scriptsize $\pm$ 0.73}& {29.16} {\scriptsize $\pm$ 0.73}\\
           \specialrule{0em}{1pt}{1pt}
           Average  & {23.96} {\scriptsize $\pm$ 0.53} & {27.04} {\scriptsize $\pm$ 0.51} & {23.79} {\scriptsize $\pm$ 0.48}& {27.49} {\scriptsize $\pm$ 0.50} & {24.53} {\scriptsize $\pm$ 0.46}& {28.00} {\scriptsize $\pm$ 0.47}\\
           \specialrule{0em}{1pt}{1pt}
			OTA~\cite{singh2020model} &  {29.10} {\scriptsize $\pm$ 0.65}  & {34.33} {\scriptsize $\pm$ 0.67} & {24.22}  {\scriptsize $\pm$ 0.53} & {27.22} {\scriptsize $\pm$ 0.59} & {24.23} {\scriptsize $\pm$ 0.60}& {25.42} {\scriptsize $\pm$ 0.63} \\
    		\specialrule{0em}{1pt}{1pt}
			DRO~\cite{wang2022meta}  & {30.43} {\scriptsize $\pm$ 0.43} & {36.21} {\scriptsize $\pm$ 0.51} & {27.56} {\scriptsize $\pm$ 0.48} & {30.19} {\scriptsize $\pm$ 0.43} & {28.33} {\scriptsize $\pm$ 0.69} & {31.24} {\scriptsize $\pm$ 0.76}\\
            \specialrule{0em}{1pt}{1pt}
            \midrule
            PURER~\cite{hu2023architecture}  & {38.66} {\scriptsize $\pm$ 0.78} & {51.95} {\scriptsize $\pm$ 0.79}& {31.14} {\scriptsize $\pm$ 0.63} &  {40.86} {\scriptsize $\pm$ 0.64}& {30.08} {\scriptsize $\pm$ 0.59} &  {40.93} {\scriptsize $\pm$ 0.66}\\
			\specialrule{0em}{1pt}{1pt}
            BiDf-MKD~\cite{hu2023learning}  & {37.66} {\scriptsize $\pm$ 0.75} & {51.16} {\scriptsize $\pm$ 0.79} & {30.66} {\scriptsize $\pm$ 0.59} & {42.30} {\scriptsize $\pm$ 0.64} & {31.62} {\scriptsize $\pm$ 0.60} & {44.32} {\scriptsize $\pm$ 0.69}\\
            \specialrule{0em}{1pt}{1pt}
            \midrule
           \textbf{Ours}   &   \bf {43.58 {\scriptsize $\pm$ 0.82}} & \bf {57.49 {\scriptsize  $\pm$ 0.79}}& \bf {36.20 {\scriptsize  $\pm$ 0.71}} & \bf {49.00 {\scriptsize  $\pm$ 0.71}}&   \bf{33.87 {\scriptsize $\pm$ 0.63}} & \bf {47.37 {\scriptsize  $\pm$ 0.73}}\\ 
           \bottomrule
	\end{tabular}}
\vspace{-1.5em}
\label{table:mainres}
\end{table*}

\begin{table}[tb]
\footnotesize
\centering
\renewcommand\arraystretch{1.1}
\setlength{\tabcolsep}{4.85pt}
\caption{\small Compare to baselines in a multi-domain scenario.}
\resizebox{0.85\linewidth}{!}{
\begin{tabular}{lcc}
			\addlinespace
			\toprule
			\specialrule{0em}{1pt}{1pt}
			\multirow{2}{*}{ \bf Method} &\multicolumn{2}{c}{\emph {CIFAR-FS}/\emph {miniImageNet}/\emph {CUB}}\\ 
			\cmidrule(l){2-3} 
			\specialrule{0em}{1pt}{1pt}
			&{$5$-way $1$-shot} & {$5$-way $5$-shot}\\
			\midrule
			\specialrule{0em}{1pt}{1pt}
			Finetune  & {24.85} {\scriptsize $\pm$ 0.54} & {28.35} {\scriptsize $\pm$ 0.61}\\
			\specialrule{0em}{1pt}{1pt}
            PURER~\cite{hu2023architecture}  & {29.67} {\scriptsize $\pm$ 0.59} & {37.96} {\scriptsize $\pm$ 0.60}\\
			\specialrule{0em}{1pt}{1pt}
            BiDf-MKD~\cite{hu2023learning}  & {31.44} {\scriptsize $\pm$ 0.64} & {40.96} {\scriptsize $\pm$ 0.59}\\
            \specialrule{0em}{1pt}{1pt}
            \midrule
           \textbf{Ours}   &   \bf {32.04 {\scriptsize $\pm$ 0.63}} & \bf {44.88 {\scriptsize  $\pm$ 0.65}}\\ 
           \bottomrule
	\end{tabular}}
\vspace{-1em}
\label{table:mainres2}
\end{table}

\subsection{Experimental Setup}
\textbf{Datasets and pre-trained models.}
We conduct experiments on two widely-used DFML benchmark datasets, and one fine-grained dataset, including \emph{miniImageNet}~\cite{miniimagenet}, \emph{CIFAR-FS}~\cite{cifarfs} and \emph{CUB}~\cite{WahCUB200_2011}. Following standard splits, we split each dataset into the meta-training, meta-validating and meta-testing subsets with disjoint label spaces.
In the DFML setting, we have no access to the meta-training data. Following~\citet{wang2022meta,hu2023learning} for a fair comparison, we collect 100 models pre-trained on 100 $N$-way tasks sampled from the meta-training subset, and those models are used as the meta-training resources.

\textbf{Implementation details.}
For model architecture, we use Conv4 as the meta-model's backbone, a standard choice in meta-learning~\cite{wu2023adaptive}.
The architecture of pre-trained models includes Conv4, ResNet-10, and ResNet-18. If not specifically stated, Conv4 is the default architecture. We use an off-the-shelf ResNet-34 pre-trained on ImageNet as our shared pre-trained model to compute the FIM.
For hyperparameters, we configure the number of task groups $c$ to 5. The step size $\beta$ for the displacement is set to 0.001, and the size of minibatches $m$ is set to 4. The memory bank $\mathcal{B}$ is limited to store 20 tasks. We report the average accuracy over 600 meta-testing tasks. \cref{sec:details} details other setups.

\textbf{Compared baselines.}
\textbf{(i) Finetune.} Train a classifier from scratch using the support set for each meta-testing task. \textbf{(ii) Best-Model.} We select the pre-trained model with the highest reported accuracy to predict the query set. \textbf{(iii) Average.} Average all pre-trained models and then finetune it using the support set. \textbf{(iv) OTA}~\cite{singh2020model}. Compute the weighted average of all pre-trained models and then fine-tune using the support set. \textbf{(v) DRO}~\cite{wang2022meta}. Meta-learn a hyper-network to fuse all pre-trained models into one single model, which serves as the meta-initialization and can be adapted to each meta-testing task using the support set. \textbf{(vi) PURER} \citep{hu2023architecture}. Adversarially train the meta-model with a learnable dataset, where a batch of pseudo tasks is sampled for meta-training at each iteration. \textbf{(vii) BiDf-MKD}~\cite{hu2023learning}. A bi-level data-free knowledge distillation framework implemented in the white-box setting to transfer meta knowledge.

\subsection{Main Results}

\paragraph{Comparisons with baselines.}
\cref{table:mainres} presents the results for few-shot classification compared with baselines. To evaluate the effectiveness of our approach, we compare it against two categories of DFML algorithm: the non-inversion methods, which enable model fusion in the parameter space, and the inversion-based methods that generate pseudo data for training. As indicated in \cref{table:mainres}, we significantly outperform all other methods.
For 1-shot learning, we outperform the leading baseline BiDf-MKD by 5.92\%, 5.54\% and 2.25\% on three datasets. For 5-shot learning, we achieve an improvement of 6.33\%, 6.70\% and 3.05\%, respectively.

\begin{table}[tb]
\footnotesize
\centering
\renewcommand\arraystretch{1.1}
\setlength{\tabcolsep}{4.85pt}
\caption{\small Compare to baselines in a multi-architecture scenario.}
\resizebox{0.85\linewidth}{!}{
\begin{tabular}{lcc}
			\addlinespace
			\toprule
			\specialrule{0em}{1pt}{1pt}
			\multirow{2}{*}{ \bf Method} &\multicolumn{2}{c}{Conv4/ResNet-10/ResNet-18}\\ 
			\cmidrule(l){2-3} 
			\specialrule{0em}{1pt}{1pt}
			&{$5$-way $1$-shot} & {$5$-way $5$-shot}\\
			\midrule
            \specialrule{0em}{1pt}{1pt}
			Finetune  & {28.59} {\scriptsize $\pm$ 0.56} & {34.77} {\scriptsize $\pm$ 0.62}\\
             \specialrule{0em}{1pt}{1pt}
            PURER~\cite{hu2023architecture}  & {39.15} {\scriptsize $\pm$ 0.70} & {49.08} {\scriptsize $\pm$ 0.74}\\
			\specialrule{0em}{1pt}{1pt}
            BiDf-MKD~\cite{hu2023learning}  & {38.08} {\scriptsize $\pm$  0.80} & {50.58} {\scriptsize $\pm$ 0.81}\\
            \specialrule{0em}{1pt}{1pt}
            \midrule
           \textbf{Ours}   &   \bf {44.18 {\scriptsize $\pm$ 0.87}} & \bf {57.15 {\scriptsize  $\pm$ 0.81}}\\ 
           \bottomrule
	\end{tabular}}
\vspace{-1em}
\label{table:mainres3}
\end{table}

\paragraph{Multi-domain scenario.}
We carry out experiments under a challenging multi-domain scenario, where each pre-trained model is designed to address different tasks from \emph{CIFAR-FS}, \emph{miniImageNet}, and \emph{CUB}. For meta-testing, the meta-model is evaluated on unseen tasks that span across three datasets. Our results are shown in \cref{table:mainres2}, outperforming the baseline by 7.19\% and 16.53\% in 1-shot and 5-shot learning, respectively. This demonstrates that our approach effectively aligns heterogeneous models from different domains, optimizing them towards a unified direction, thereby achieving robust generalization across various domains.

\begin{table*}[tb]
  \footnotesize
    \centering
    \renewcommand\arraystretch{1.1}
    \setlength{\tabcolsep}{4.85pt}
  \caption{Effects of the proposed modules.}
   \resizebox{0.82\linewidth}{!}{
  \begin{tabular}{lcccccccc}
    \toprule
    \multirow{2}{*}{\textbf{Setting}} & \multicolumn{2}{c}{Module} & \multicolumn{2}{c}{\emph{CIFAR-FS}} & \multicolumn{2}{c}{\emph {miniImageNet}} & \multicolumn{2}{c}{\emph {CUB}}\\
    \cmidrule(r){2-3}
    \cmidrule(r){4-5}
    \cmidrule(r){6-7}
    \cmidrule(r){8-9}
    &\textbf{Group}&\textbf{IGR}&{5-way 1-shot}&{5-way 5-shot}&{5-way 1-shot}&{5-way 5-shot}&{5-way 1-shot}&{5-way 5-shot}\\
    \midrule
    Vanilla&&&41.84 {\scriptsize $\pm$ 0.82}&55.01 {\scriptsize $\pm$ 0.77}&34.96 {\scriptsize $\pm$ 0.70}	&47.24 {\scriptsize $\pm$ 0.69}	&32.28 {\scriptsize $\pm$ 0.65}	&44.82 {\scriptsize $\pm$ 0.67}\\
    \midrule
    &\checkmark&& 43.06 {\scriptsize $\pm$ 0.81} &55.84 {\scriptsize $\pm$ 0.76} &36.01 {\scriptsize $\pm$ 0.70}	&48.53 {\scriptsize $\pm$ 0.69}	&32.88 {\scriptsize $\pm$ 0.60}	&47.25 {\scriptsize $\pm$ 0.66}\\
    &&\checkmark&43.25 {\scriptsize $\pm$ 0.85}&55.54 {\scriptsize $\pm$ 0.74}&\bf{36.78 {\scriptsize $\pm$ 0.70}	}&48.35 {\scriptsize $\pm$ 0.71}	&33.06 {\scriptsize $\pm$ 0.64}	&\bf{47.92 {\scriptsize $\pm$ 0.69}}\\
    \midrule
    \textbf{Ours}&\checkmark&\checkmark&\bf {43.58 {\scriptsize $\pm$ 0.82}}&\bf {57.49 {\scriptsize $\pm$ 0.79}}&{36.20 {\scriptsize $\pm$ 0.71}}	&\bf {49.00 {\scriptsize $\pm$ 0.71}}	&\bf {33.87 {\scriptsize $\pm$ 0.63}}	& {47.37 {\scriptsize $\pm$ 0.73}}\\
     \bottomrule
  \end{tabular}}
  \label{tab:ablation}
  \vspace{-1.5em}
\end{table*}

\paragraph{Multi-architecture scenario.}
We also conduct experiments in a multi-architecture scenario where each pre-trained model differs in architecture. For each pre-trained model, the architecture is randomly chosen from Conv4, ResNet-10, and ResNet-18. The results are presented in \cref{table:mainres3}. It outperforms all baselines and can apply to multi-architecture scenario without any change. This flexibility is attributable to our method's lack of constraints on the underlying architecture or scale of the pre-trained models, which enables effective alignment across heterogeneous models.

\subsection{Ablation Studies}
\paragraph{Effects of each module.}
\cref{tab:ablation} assesses the contribution of each module to performance. Initially, we introduce the vanilla only performing meta-model optimization. The vanilla still manages to outperform the baselines mentioned in \cref{table:mainres}, indicating its capacity for rapid adaptation to unseen tasks. When we integrate task grouping into the approach, we observe significant improvements. This indicates that training heterogeneous models together enables the meta-model to learn shared representations across tasks. However, this method (\ie, ERM) merely aggregates losses from multiple tasks to update the meta-model, without aligning the optimization directions across these tasks. By adding implicit gradient regularization, we observe a significant improvement, demonstrating the effectiveness of the conflicting task regularization. With all modules, we achieve the best performance with a boosting improvement of 2\%-3\%, showing their complementarity.

\begin{figure}[tb]
\centering
\begin{minipage}[t]{\linewidth}
        \centering
        \subfloat{\includegraphics[width=0.49\columnwidth]{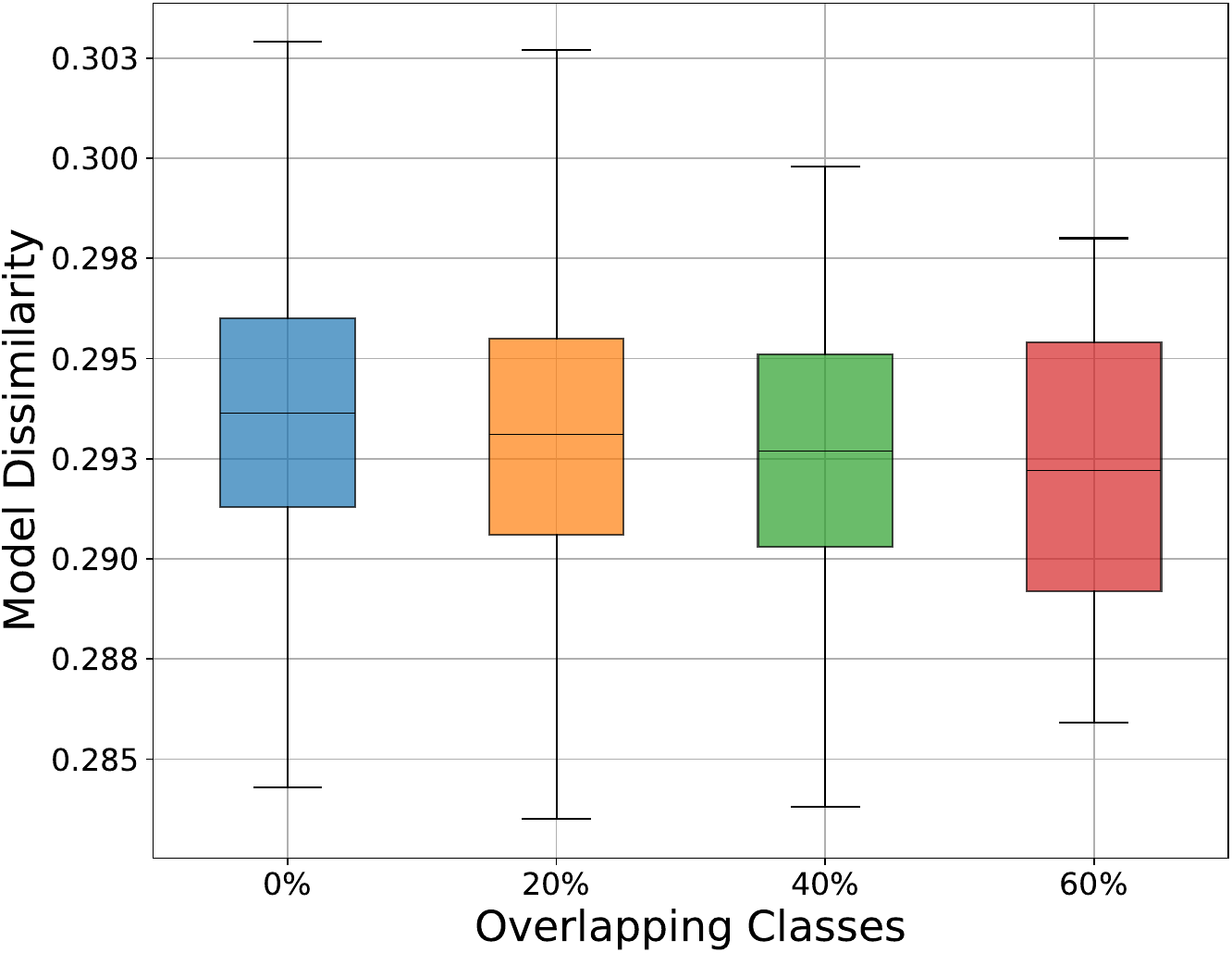} }
        \subfloat{\includegraphics[width=0.49\columnwidth]{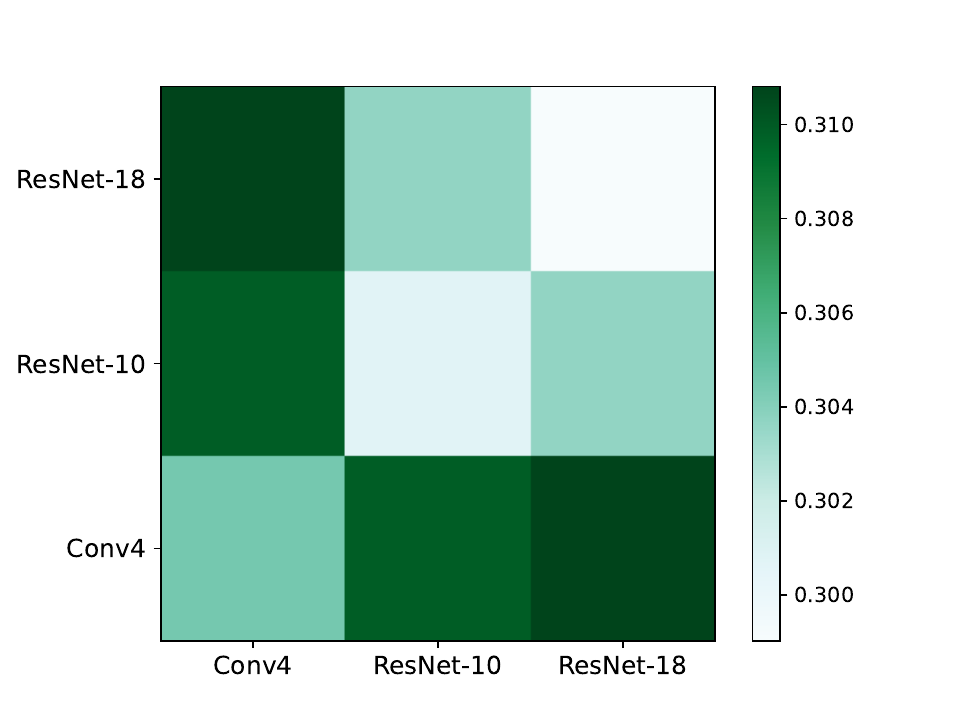} }
        \vspace{-0.5em}
         \caption{\small Model dissimilarity measured by FIM accurately reflects the model heterogeneity. We assess model heterogeneity by overlapping training classes and distinct model architectures.}
\label{fig:overlap}
\end{minipage}
\vspace{-1em}
\end{figure}

\paragraph{Model dissimilarity measurement.}
We also confirm that our model dissimilarity measurement accurately reflects model heterogeneity, facilitating the grouping of pre-trained models. We assess the model heterogeneity using the ratio of overlapping classes and model architectures. The premise is that fewer overlapping training classes or differing architectures imply higher model heterogeneity. We conduct experiments with 100 pre-trained Conv4 models, as shown in \cref{fig:overlap} (left). There is a negative correlation between the ratio of overlapping classes and the model dissimilarity. In \cref{fig:overlap} (right), we select 100 each of Conv4, ResNet-10, and ResNet-18. Results indicate that models with different architectures exhibit higher dissimilarity, with the greatest dissimilarity observed between Conv4 and ResNet-18.

\begin{figure}[tb]
\begin{minipage}[t]{\linewidth}
        \centering
        \subfloat[Gradient regularizer loss]{\includegraphics[width=0.49\columnwidth]{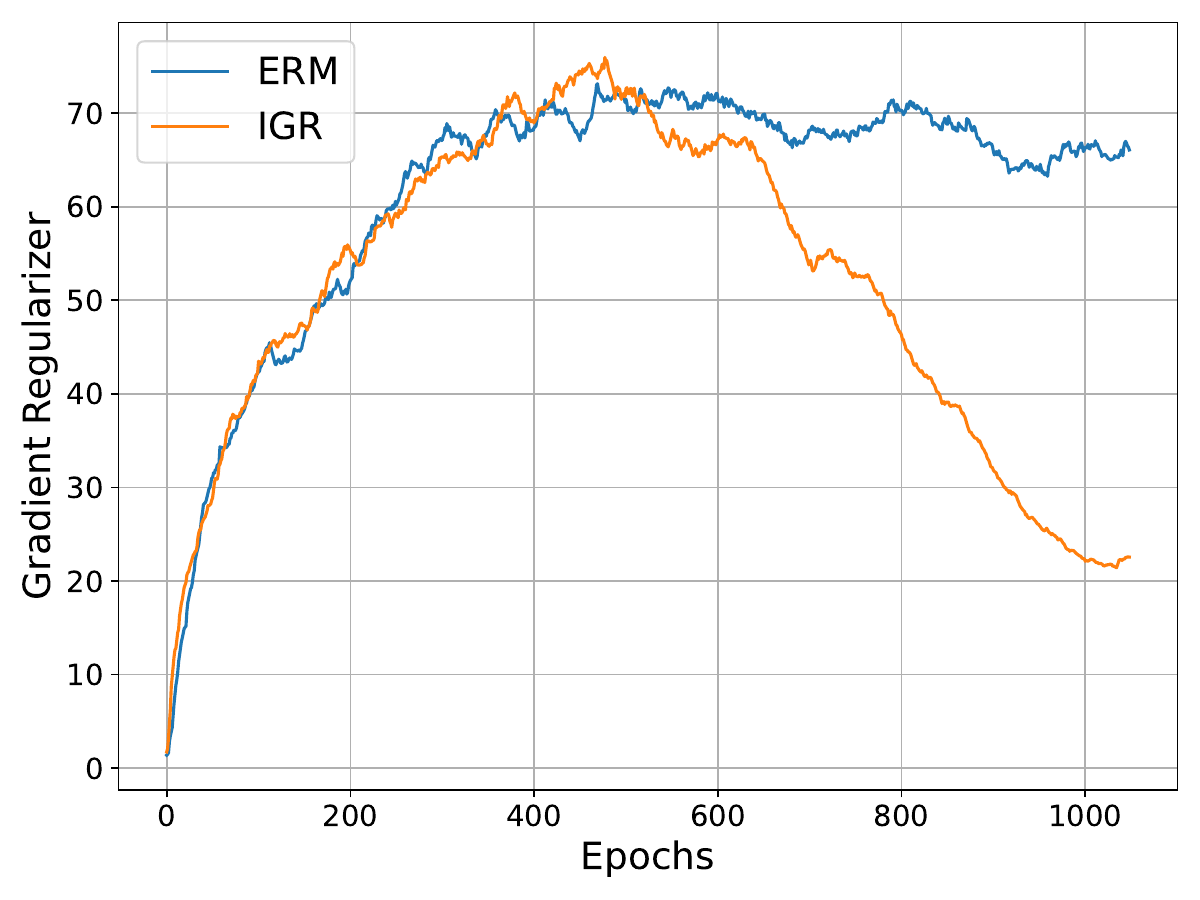} }
        \subfloat[Gradient cosine similarity]{\includegraphics[width=0.49\columnwidth]{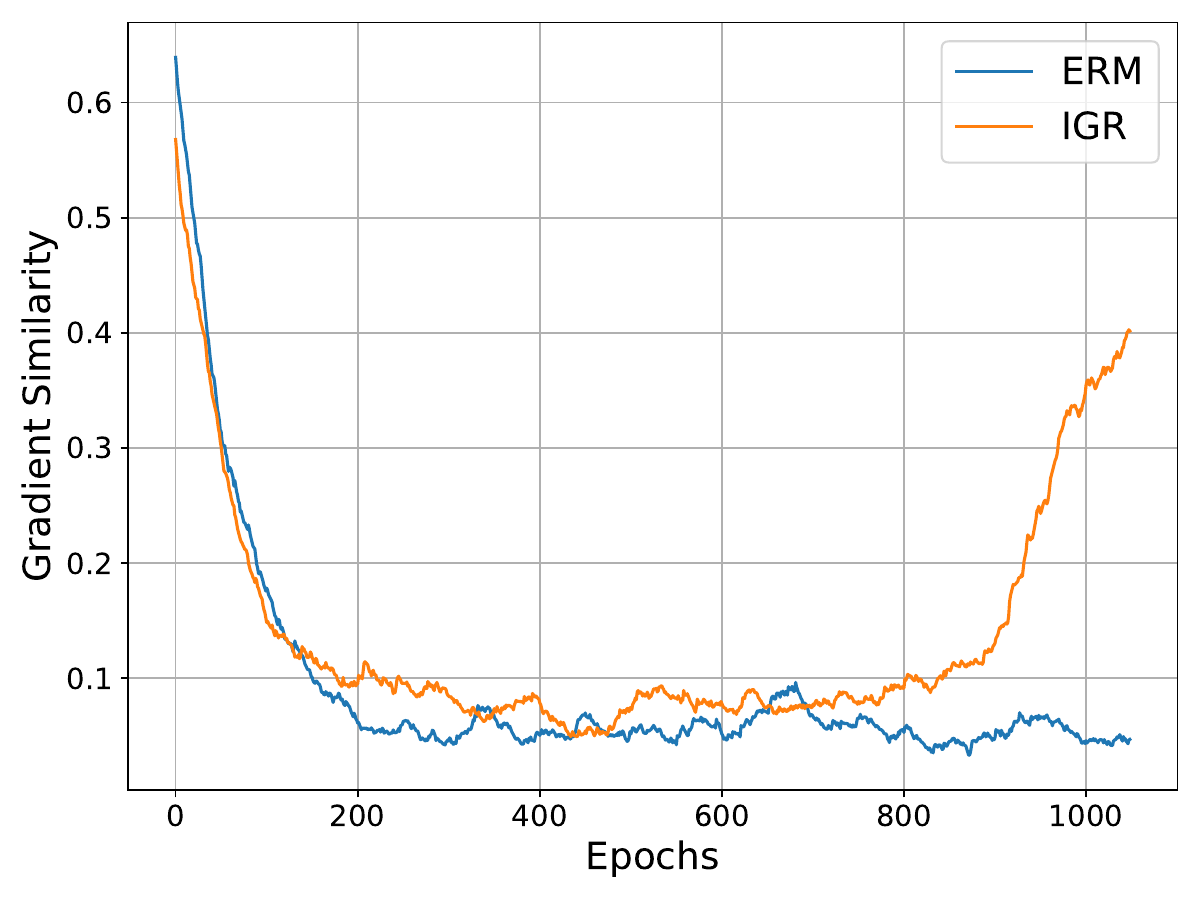} }
        \vspace{-0.7em}
         \caption{\small The gradient discrepancy across tasks. We plot the progression of gradient regularizer loss and gradient cosine similarity during training, confirming the efficacy of IGR in implicitly minimizing the gradient regularizer and aligning gradient directions.}
	\label{fig:gdm}
\end{minipage}
\vspace{-1em}
\end{figure}

\begin{table*}[tb]
\footnotesize
\centering
\renewcommand\arraystretch{1.1}
\setlength{\tabcolsep}{4.85pt}
\caption{\small Ablation studies on the four loss components.}
\resizebox{0.82\linewidth}{!}{
\begin{tabular}{lcccccc}
			\addlinespace
			\toprule
			\specialrule{0em}{1pt}{1pt}
			\multirow{2}{*}{ \bf Method} &\multicolumn{2}{c}{\emph {CIFAR-FS}} & \multicolumn{2}{c}{\emph {miniImageNet}} & \multicolumn{2}{c}{\emph {CUB}}\\ 
			\cmidrule(l){2-3} \cmidrule(l){4-5} \cmidrule(l){6-7}
			\specialrule{0em}{1pt}{1pt}
			&{$5$-way $1$-shot} & {$5$-way $5$-shot} &{$5$-way $1$-shot} &{$5$-way $5$-shot} &{$5$-way $1$-shot} &{$5$-way $5$-shot}\\
			\midrule
			\specialrule{0em}{1pt}{1pt}
			w/o $\mathcal{L}_ {CE}$  & {32.36} {\scriptsize $\pm$ 0.59} & {46.22} {\scriptsize $\pm$ 0.73} &{31.43} {\scriptsize $\pm$ 0.56}& {44.36} {\scriptsize $\pm$ 0.65}  &{28.20} {\scriptsize $\pm$ 0.55}& {36.58} {\scriptsize $\pm$ 0.58}\\
			\specialrule{0em}{1pt}{1pt}
           w/o $\mathcal{L}_ {BN}$	  & {41.28} {\scriptsize $\pm$ 0.80} & {54.71} {\scriptsize $\pm$ 0.76} & {34.18} {\scriptsize $\pm$ 0.59}& {45.34} {\scriptsize $\pm$ 0.64} & {33.41} {\scriptsize $\pm$ 0.61}& {47.29} {\scriptsize $\pm$ 0.69}\\
           \specialrule{0em}{1pt}{1pt}
           w/o soft labels  & {42.87} {\scriptsize $\pm$ 0.84} & {55.53} {\scriptsize $\pm$ 0.78} & {36.03} {\scriptsize $\pm$ 0.70}& {47.49} {\scriptsize $\pm$ 0.74} & {32.71} {\scriptsize $\pm$ 0.64}& {46.74} {\scriptsize $\pm$ 0.69}\\
           \specialrule{0em}{1pt}{1pt}
		w/o hard labels &  {43.54} {\scriptsize $\pm$ 0.82}  & {55.75} {\scriptsize $\pm$ 0.77} & {35.70}  {\scriptsize $\pm$ 0.70} & {47.96} {\scriptsize $\pm$ 0.74} & {33.52} {\scriptsize $\pm$ 0.62}& \bf{47.72 {\scriptsize $\pm$ 0.70}} \\
    		\specialrule{0em}{1pt}{1pt}
            \midrule
           w/ all components   &   \bf {43.58 {\scriptsize $\pm$ 0.82}} & \bf {57.49 {\scriptsize  $\pm$ 0.79}}& \bf {36.20 {\scriptsize  $\pm$ 0.71}} & \bf {49.00 {\scriptsize  $\pm$ 0.71}}&   \bf{33.87 {\scriptsize $\pm$ 0.63}} &  {47.37 {\scriptsize  $\pm$ 0.73}}\\ 
           \bottomrule
	\end{tabular}}
\vspace{-1.8em}
\label{table:ablloss}
\end{table*}

\paragraph{Gradient discrepancy minimization.}
We incorporate implicit gradient regularization (IGR) to align gradients across different tasks, steering the meta-model in a unified direction. To ascertain whether our approach indeed minimizes gradient discrepancies, we plot the progression of gradient regularizer loss and gradient cosine similarity during training with different objectives, as depicted in \cref{fig:gdm}. We train both IGR (blue) and ERM (orange) on \emph{CIFAR-FS} until convergence, while tracking the gradient regularizer $\frac{1}{2m}\sum_{i=0}^{m-1}||\nabla{\mathcal{L}_i}(\boldsymbol{\theta})-\nabla \bar {\mathcal{L}}(\boldsymbol{\theta})||^2$ across tasks from heterogeneous pre-trained models. To ensure a fair comparison, we use the exact same sequence of tasks
for IGR and ERM.

As shown in \cref{fig:gdm}, the gradient regularizer of IGR gradually decreases with ongoing training. Conversely, ERM maintains high values until the end. Moreover, we evaluate gradient cosine similarity, which measures the angle between gradients of two tasks. It is clear that during training, the gradient similarity of IGR increases, whereas it remains constant for ERM. These observations confirm the efficacy of IGR in implicitly minimizing the gradient regularizer and aligning gradient directions.

\begin{figure}[tb]
\centering
\begin{minipage}[t]{\linewidth}
        \subfloat[5-way 1-shot]{\includegraphics[width=0.49\columnwidth]{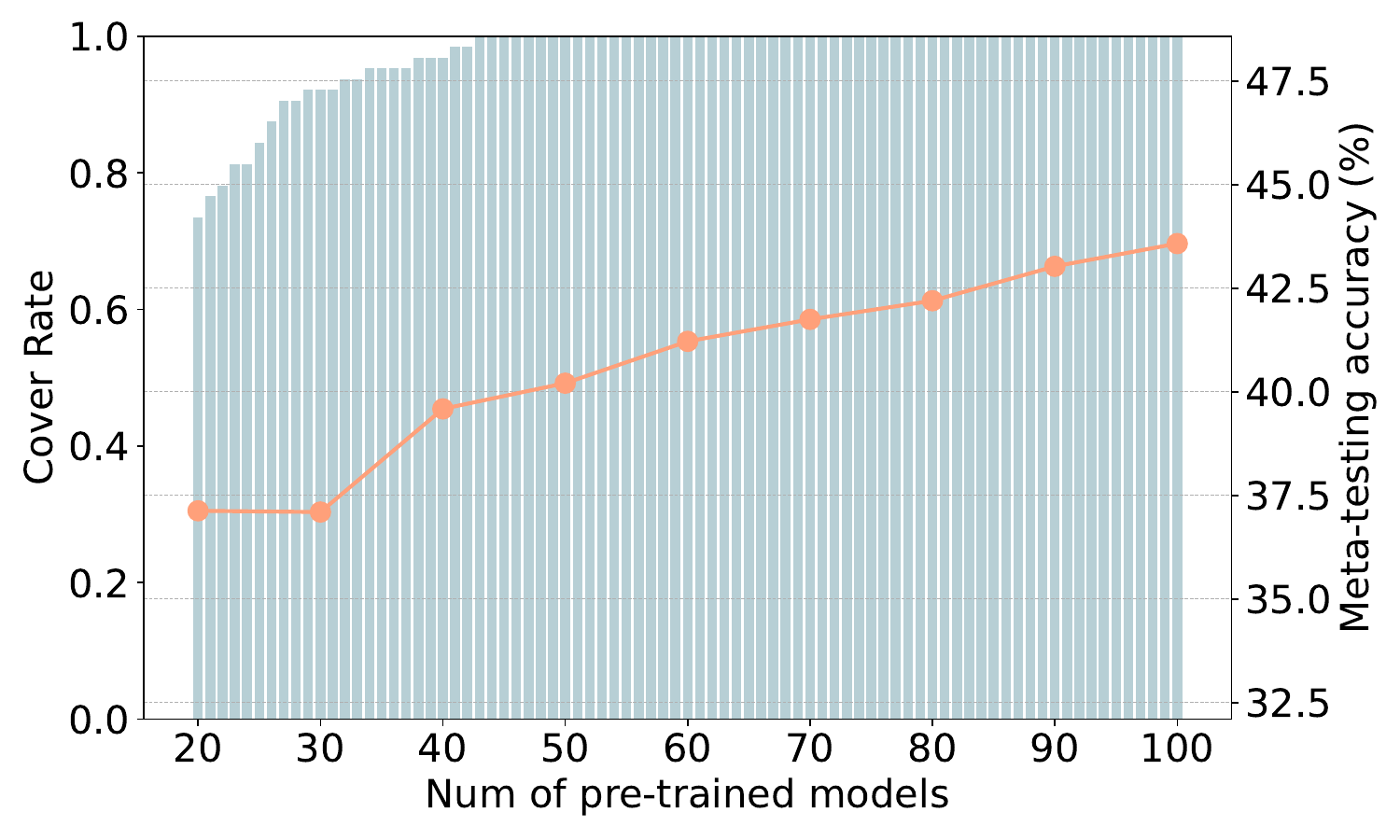} }
        \subfloat[5-way 5-shot]{\includegraphics[width=0.49\columnwidth]{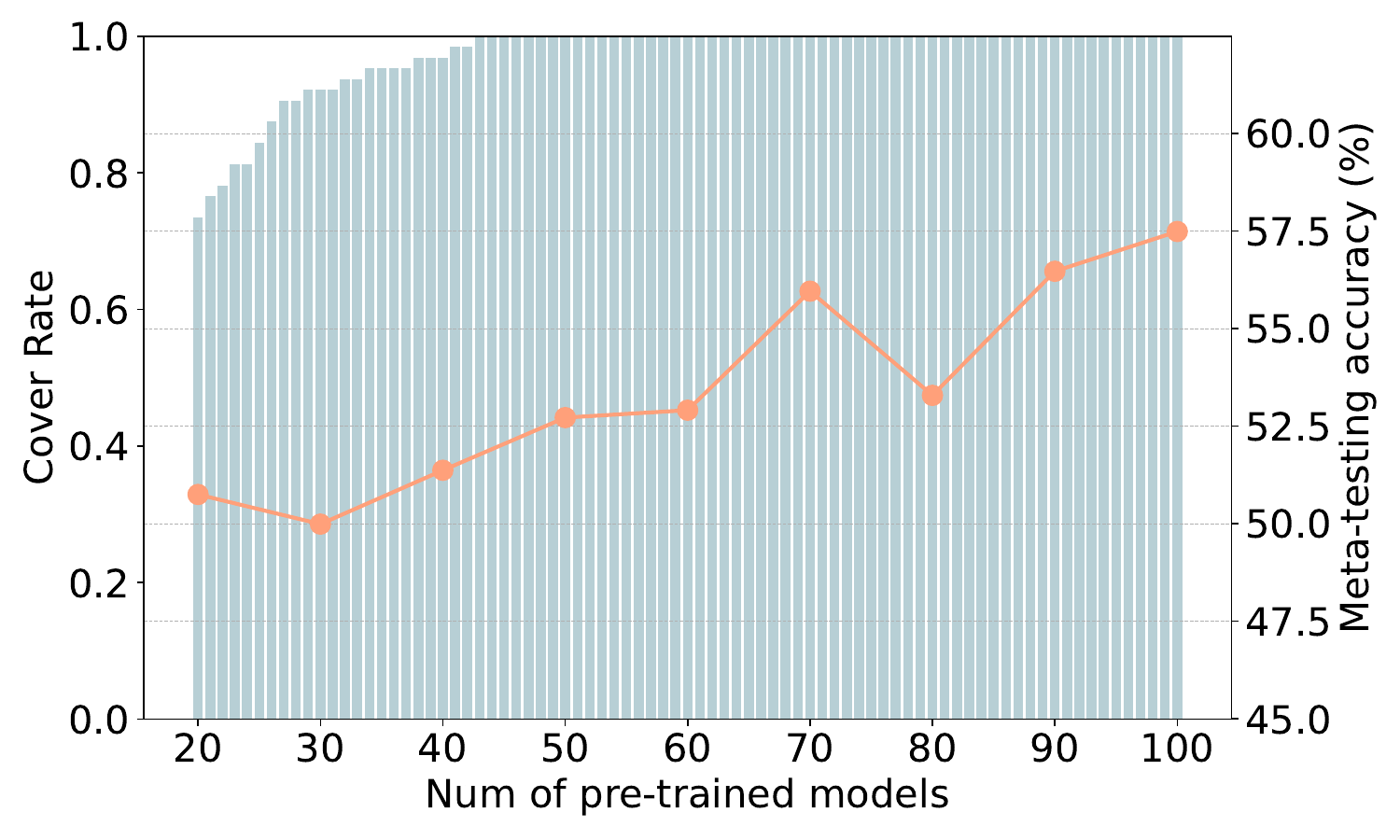} }
        \vspace{-0.5em}
         \caption{The relationship between the accuracy and the number of pre-trained models.}
	\label{fig:number}
\end{minipage}
\vspace{-1em}
\end{figure}

\paragraph{Effects of the number of pre-trained models.}
We carry out experiments with varying numbers of pre-trained models, each individually pre-trained on a 5-way subset of \emph{CIFAR-FS}. Our method demonstrates scalability to any number of pre-trained models, with a consistent improvement in performance as more models are added (refer to \cref{fig:number}). We also introduce an intrinsic factor termed the ``cover rate'', which reflects the coverage rate of classes within the meta-training subset. As the number of pre-trained models grows, so does the cover rate. Notably, even when the cover rate reaches 100\% (which occurs with more than 50 pre-trained models), we continue to observe performance enhancements. This improvement is attributed to the increased diversity of pre-trained models, providing a more comprehensive understanding of various classes. This, in turn, strengthens the generalization ability of the meta-model when encountering previously unseen classes.

\paragraph{Effects of the number of task groupings.}
For task grouping algorithms, we conduct experiments with varying numbers of task groupings. As shown in Table~\ref{table:number}, performance initially improves with an increasing number of groupings, while it diminishes when the number of groupings becomes overly large. This is because a proper number of groupings enables learning of shared representations among heterogeneous models. However, too many groups result in a decreased number of tasks per group, thereby reducing the benefits derived from inter-task information sharing. In future work, we plan to explore methods for automatically selecting an appropriate number of groupings, \eg, by utilizing cluster evaluation metrics~\cite{maulik2002performance} such as the Calinski-Harabasz index.

\begin{table}[tb!]
\footnotesize
\centering
\setlength{\tabcolsep}{5.0pt}
\caption{\small Effects of the number of task groupings.}
\resizebox{\linewidth}{!}{
\begin{tabular}{cccccccccc}
\toprule
\bf{Number} & 1     & 2     & 3     & 4& 5& 6& 7& 8& 9\\\midrule
5-way 5-shot & 48.11 & 48.79 & 48.20 & 48.75& 49.00& 49.37& 49.31& 49.24& 48.02
 \\
\bottomrule                
\end{tabular}}
\vspace{-1em}
\label{table:number}
\end{table}

\paragraph{Effects of the loss components.}
The classification loss $\mathcal{L}_ {CE}$ is introduced for class-specific generation, crucial for the following meta-learning process. It prevents the generated images from becoming a blend of attributes from multiple classes. The regularization loss $\mathcal{L}_ {BN}$ utilizes Batch Normalization statistics to ensure the synthetic images' distribution aligns closer to the original images. Soft labels convey the teacher model's knowledge, utilizing semantic relationships for richer supervision. Hard labels, as the ground truth, offer precise supervisory information to prevent errors introduced by soft labels. As shown in \cref{table:ablloss}, removing $\mathcal{L}_ {CE}$, $\mathcal{L}_ {BN}$, soft labels, and hard labels results in average performance decreases of 8.06\%, 1.88\%, 1.02\%, and 0.55\%, respectively. On the \emph{CUB} dataset, hard labels even have a negative impact, possibly due to the overly strong one-hot assumption when distinguishing fine-grained categories. \cref{table:ablloss} indicate that the classification loss and soft labels loss have a considerable impact on the results.
\section{Conclusion}
Current DFML methods often overlook the heterogeneity among pre-trained models.
In this paper, we empirically and theoretically rethink the model heterogeneity in DFML. We find that the model heterogeneity introduces a heterogeneity-homogeneity trade-off.
Based on our findings, we propose Task Groupings Regularization that benefits from model heterogeneity by grouping and aligning conflicting tasks within each group, to capture shared representations that generalize across tasks.
Comprehensive experiments showcase the superiority of our approach in multiple benchmarks.
Exploring its extension to other domains, such as NLP, represents a valuable direction for future research.

\section*{Acknowledgements}
This work was supported by the National Key R\&D Program of China (2022YFB4701400/4701402), SSTIC Grant(KJZD20230923115106012), Shenzhen Key Laboratory (ZDSYS20210623092001004), and Beijing Key Lab of Networked Multimedia.

\section*{Impact Statement}
Data-Free Meta-Learning (DFML) provides an effective resolution to the ethical, privacy, and labor challenges posed by large-scale image datasets, which often result in the retraction or restriction of these resources and prompt a shift toward using pre-trained models for research. However, while DFML mitigates many of these concerns, the model inversion technique it employs could potentially expose information from the original training data, presenting a risk of data leakage. Despite this, DFML significantly benefits socially by enabling few-shot learning in data-limited environments, thus promoting substantial positive impacts in fields where data scarcity is a challenge.

\bibliography{main_paper}
\bibliographystyle{icml2024}

\newpage
\appendix
\onecolumn

\section{Implementation Details}
\label{sec:details}
The architecture for the meta-model is based on Conv4. Common in meta-learning studies~\cite{chen2020variational,liu2020adaptive}, Conv4 consists of four convolutional blocks, with each block comprising 32 $3 \times 3$ filters, a BatchNorm layer, a ReLU function, and a $2 \times 2$ max-pooling layer.
All pre-trained models are constructed to solve various 5-way tasks by randomly choosing 5 classes from the meta-training set (be it \emph{CIFAR-FS}, \emph{miniImageNet}, or \emph{CUB}). With the Adam optimizer, these models are pre-trained using standard supervised learning at a learning rate of 0.01. We collect a set of 100 pre-trained models following previous works.
For computing the FIM, we use the off-the-shelf ResNet-34\footnote{https://pytorch.org/vision/stable/models/generated/torchvision.models.resnet34} pre-trained on ImageNet as our shared model.
The total epoch $T$ is set to 1000 for \emph{CIFAR-FS} and \emph{CUB}, and 1200 for \emph{miniImageNet}. For the task recovery, we generate $\hat{\boldsymbol{X}_{i}}$ with 30 images in each epoch. The generator parameters $\boldsymbol{\theta}_{G}$ and latent code $\boldsymbol{Z}$ are optimized simultaneously using the Adam optimizer, aiming to minimize \cref{eq:Loss_g} over 200 steps with a learning rate of 0.001. Similarly, the meta-model parameters $\boldsymbol{\theta}$ are updated with gradient $\boldsymbol{g}_{IGR}$ with a learning rate set at 0.001.
For the cross-task replay, we employ MAML to execute meta-learning on tasks sampled in the memory bank. Following~\citet{hu2023architecture,hu2023learning}, we use the Adam optimizer, set with inner and outer loop learning rates of 0.01 and 0.001, respectively.

In the multi-domain scenario, pre-trained models are designed to solve different 5-way tasks, which are devised by randomly sampling 5 classes from the meta-training sets, including \emph{CIFAR-FS}, \emph{miniImageNet}, and \emph{CUB} \cite{han2023region,wei2024task}. We also take Conv4 as the architecture of pre-trained models and the meta-model. For meta-testing, we evenly construct 1800 meta-testing tasks across all meta-testing sets, including \emph{CIFAR-FS}, \emph{miniImageNet}, and \emph{CUB}.

In the multi-architecture scenario, we focus on pre-trained models with different architectures. Specifically, we consider Conv4, ResNet-10, and ResNet-18 as the diverse architectures represented within these pre-trained models. ResNet-10 and ResNet-18 are larger-scale neural networks compared to Conv4. We adopt Conv4 as the architecture for the meta-model. All pre-trained models are designed for solving diverse 5-way tasks, which are constructed by randomly sampling 5 classes from \emph{CIFAR-FS}. All other configurations remain consistent with the above settings.

\section{Architecture of the Generator}
\label{sec:generator}
The structure of the generator for task recovery is detailed in \cref{tab:sturcture_of_generator}. This generator processes standard Gaussian noise as input to produce the recovered data. The dimension of the Gaussian noise data, $\boldsymbol{z}$, is designated as 256 ($d_{\boldsymbol{z}}$). For the LeakyReLU activation, the $negative\_slope$ parameter is set to 0.2. We specify the image size to be 32 for models pre-trained on \emph{CIFAR-FS} and 84 for those pre-trained on \emph{miniImageNet} and \emph{CUB}. The number of channels ($nc$) for color image recovery is set to 3, and the number of convolutional filters ($nf$) is set to 64.
\vspace{-0.5em}
\begin{table*}[bh]\label{app:sturcture_of_generator}
    \centering
    \caption{\small Detailed structure of the generator. We highlight the dimension change in \textcolor{red}{red}.}
    \scalebox{0.7}{
    \begin{tabular}{ccc}
    \toprule
    \textbf{Notion} &\multicolumn{2}{c}{\textbf{Description}}\\
    \midrule
    $img\_size$ $\times$ $img\_size$&\multicolumn{2}{c}{ resolution of recovered image}\\
    $bs$ & \multicolumn{2}{c}{ batch size}\\
    $nc$ & \multicolumn{2}{c}{ number of channels of recovered image}\\
    $nf$ & \multicolumn{2}{c}{ number of convolutional filters}\\
    FC($\cdot$) & \multicolumn{2}{c}{ fully connected layer;}\\
    BN($\cdot$)&\multicolumn{2}{c}{ batch normalization layer}\\
    Conv2D($input$,\ $output$,$filter\_size$,\ $stride$,\ $padding$) & \multicolumn{2}{c}{ convolutional layer}\\
    \toprule
    \multirow{2}{*}{\textbf{Structure}} & \multicolumn{2}{c}{\textbf{Dimension}}\\
    & \textbf{Before} & \textbf{After}\\
     \midrule
         $\boldsymbol{z} \in \mathbb{R}_{d_{\boldsymbol{z}}} \sim \mathcal{N}(\boldsymbol{0},\boldsymbol{1})$&$\left[\ \textcolor{red}{1}, d_{\boldsymbol{z}}\ \right]$ & $\left[\ \textcolor{red}{bs}, d_{\boldsymbol{z}}\ \right]$  \\
         
        FC($\boldsymbol{Z}$) & $\left[\ bs, \textcolor{red}{d_{\boldsymbol{z}}}\ \right] $ & $ \left[\ bs, \textcolor{red}{2 \times nf \times (img\_size//4) \times (img\_size//4)}\ \right]$ \\
    
        Reshape & $\left[\ bs, \textcolor{red}{2 \times nf \times (img\_size//4) \times (img\_size//4)}\ \right] $ & $ \left[\ bs, \textcolor{red}{2 \times nf}, \textcolor{red}{(img\_size//4)}, \textcolor{red}{(img\_size//4)}\ \right]$\\
        \midrule
         BN & $ \left[\ bs, 2 \times nf, (img\_size//4), (img\_size//4)\ \right]$&$ \left[\ bs, 2 \times nf, (img\_size//4), (img\_size//4)\ \right]$\\
        Upsampling &$ \left[\ bs, 2 \times nf, (img\_size//\textcolor{red}{4}), (img\_size//\textcolor{red}{4}))\ \right]$ & $ \left[\ bs, 2 \times nf, (img\_size//\textcolor{red}{2}), (img\_size//\textcolor{red}{2}))\ \right]$\\
        \midrule
        Conv2D($2 \times nf,\ 2 \times nf,\ 3,\ 1,\ 1$) & $ \left[\ bs, 2 \times nf, (img\_size//2), (img\_size//2))\ \right]$ & $ \left[\ bs, 2 \times nf, (img\_size//2), (img\_size//2))\ \right]$\\
        BN, LeakyReLU & $ \left[\ bs, 2 \times nf, (img\_size//2), (img\_size//2))\ \right]$&$ \left[\ bs, 2 \times nf, (img\_size//2), (img\_size//2))\ \right]$ \\
        Upsampling& $ \left[\ bs, 2 \times nf, (img\_size//\textcolor{red}{2}), (img\_size//\textcolor{red}{2}))\ \right]$ & $ \left[\ bs, 2 \times nf, img\_size, img\_size\ \right]$\\
        \midrule
        Conv2D($2 \times nf,\ nf,\ 3,\ 1,\ 1$) & $ \left[\ bs, \textcolor{red}{2 \times nf}, img\_size, img\_size\ \right]$ & $ \left[\ bs, \textcolor{red}{nf}, img\_size, img\_size\ \right]$\\
        BN, LeakyReLU& $ \left[\ bs, nf, img\_size, img\_size\ \right]$ & $ \left[\ bs, nf, img\_size, img\_size\ \right]$\\
        Conv2D($nf,\ nc,\ 3,\ 1,\ 1$) & $ \left[\ bs, \textcolor{red}{nf}, img\_size, img\_size\ \right]$ & $ \left[\ bs, \textcolor{red}{nc}, img\_size, img\_size\ \right]$\\
        Sigmoid & $ \left[\ bs, nc, img\_size, img\_size\ \right]$& $ \left[\ bs, nc, img\_size, img\_size\ \right]$\\
        \bottomrule
    \end{tabular}
    }
    \label{tab:sturcture_of_generator}
\end{table*}

\section{Theoretical Proofs}
\label{sec:proof}

\begin{theorem}\label{app:theorem0}
Assume $M_{meta}(\cdot;\boldsymbol{\theta})$ is probably approximately correct (PAC), \ie, there exists $\zeta(N,\delta)\geq0$ monotonically decreasing with $N$, and the loss function $\ell(\cdot)$ is $K$-Lipschitz continuous. Then, with probability at least $1-2 \delta$ the following bounds hold:
\begin{align*}
&|E(\boldsymbol{\theta}(M_\mathrm{bas}),\boldsymbol{\theta}(M_\mathrm{aux}))-
\hat{E}(\boldsymbol{\theta}(M_\mathrm{bas},M_\mathrm{aux}))|\\
\leq &\zeta(N,\delta)
+K\sum_{t=\mathrm{bas}}^{\mathrm{aux}}\mathbb{E}_{\mathcal{P}_t}[|M_{meta}(\boldsymbol{\theta}(M_t))-M_{meta}(\boldsymbol{\theta}(M_\mathrm{bas},M_\mathrm{aux}))|], 
\end{align*}
where $E=\mathbb{E}_{\mathcal{P}_\mathrm{bas}}[\ell\left(M_{meta}(x,\boldsymbol{\theta}(M_\mathrm{bas}))\right)]+\mathbb{E}_{\mathcal{P}_\mathrm{aux}}[\ell\left(M_{meta}(x,\boldsymbol{\theta}(M_\mathrm{aux}))\right)]$ is the expected error over the merged distribution and $\hat{E}=\frac{1}{N}\sum_{i=1}^{N}\ell\left(M_{meta}(x_i,\boldsymbol{\theta}(M_\mathrm{bas},M_\mathrm{aux}))\right)$ is the empirical error over the $N$ training samples.
\end{theorem}

\begin{proof}
Using PAC-Bayesian generalization bound~\cite{mcallester1999pac}, for all $\boldsymbol{\theta}\in\boldsymbol{\Theta}$, with probability $1-\delta$ over independent draws $x_{i} \sim \mathcal{P}_t$, we have:
\begin{align*}
\left|\mathbb{E}_{\mathcal{P}_t}[\ell(M_{meta}(x,\boldsymbol{\theta}))]-\frac{1}{N}\sum_{i=1}^{N}\ell(M_{meta}(x_i,\boldsymbol{\theta}))\right|\leq\zeta(N,\delta)=\sqrt{\frac{\log\left|\boldsymbol{\Theta}\right|+\log\frac 1\delta}{2N}}.
\end{align*}

This comes as a generalization of the law of large numbers for the case in which samples are \emph{i.i.d.}, where the error is of order $1/N$. If the loss function $\ell(\cdot)$ is $K$-Lipschitz continuous, we have:
\begin{align*}
|\ell(M_{meta}(x,\boldsymbol{\theta}_1))-\ell(M_{meta}(x,\boldsymbol{\theta}_2))|\leq K\left|M_{meta}(x,\boldsymbol{\theta}_1)-M_{meta}(x,\boldsymbol{\theta}_2)\right|.
\end{align*}

By a direct application of the triangle inequality and the PAC bound over the merged probability distribution, \ie, $\bar{\mathcal{P}}=\mathcal{P}_\mathrm{bas}+\mathcal{P}_\mathrm{aux}$, and taking the Lipschitz assumption over the solutions, then:

\begin{align*}
&\quad\,\,|E(\boldsymbol{\theta}(M_\mathrm{bas}),\boldsymbol{\theta}(M_\mathrm{aux}))-
\hat{E}(\boldsymbol{\theta}(M_\mathrm{bas},M_\mathrm{aux}))|\\
&=|\mathbb{E}_{\mathcal{P}_\mathrm{bas}}[\ell\left(M_{meta}(x,\boldsymbol{\theta}(M_\mathrm{bas}))\right)]+\mathbb{E}_{\mathcal{P}_\mathrm{aux}}[\ell\left(M_{meta}(x,\boldsymbol{\theta}(M_\mathrm{aux}))\right)]
-\frac{1}{N}\sum_{i=1}^{N}\ell\left(M_{meta}(x_i,\boldsymbol{\theta}(M_\mathrm{bas},M_\mathrm{aux}))\right)|\\
&\leq \zeta(N,\delta) + K\mathbb{E}_{\mathcal{P}_\mathrm{bas}}[|M_{meta}(x,\boldsymbol{\theta}(M_\mathrm{bas}))-M_{meta}(x,\boldsymbol{\theta}(M_\mathrm{bas},M_\mathrm{aux}))|]\\
&\quad+K\mathbb{E}_{\mathcal{P}_\mathrm{aux}}[|M_{meta}(x,\boldsymbol{\theta}(M_\mathrm{aux}))-M_{meta}(x,\boldsymbol{\theta}(M_\mathrm{bas},M_\mathrm{aux}))|]\\
&= \zeta(N,\delta) + K\sum_{t=\mathrm{bas}}^{\mathrm{aux}}\mathbb{E}_{\mathcal{P}_t}[|M_{meta}(\boldsymbol{\theta}(M_t))-M_{meta}(\boldsymbol{\theta}(M_\mathrm{bas},M_\mathrm{aux}))|].
\end{align*}

Here, we choose the notation $M_{meta}(x,\boldsymbol{\theta}(\cdot)) = M_{meta}(\boldsymbol{\theta}(\cdot))$ for brevity, referring to the prediction function for inputs $x$.

\end{proof}

\begin{theorem}\label{app:theorem1}
If ${\mathcal{L}_i}(\boldsymbol{\theta})$ has Lipschitz Hessian, then the update gradient $\boldsymbol{g}_{IGR}=\frac{1}{m}\sum_{i=0}^{m-1}\nabla {\mathcal{L}_i}(\boldsymbol{\theta}-\boldsymbol{v}_i(\boldsymbol{\theta}))$, calculated following an initial displacement applied to the parameters $\boldsymbol{\theta}$, can be expressed as follows:
\begin{align*}
\boldsymbol{g}_{IGR}&=\nabla \bar {\mathcal{L}}(\boldsymbol{\theta}) + \underbrace{\frac{\beta}{2m}\nabla(\sum_{i=0}^{m-1}\|\nabla \mathcal{L}_i(\boldsymbol{\theta})-\nabla \bar {\mathcal{L}}(\boldsymbol{\theta})\|^2)}_{Gradient\,Regularization}+\mathcal{O}(\beta^2),
\end{align*}
\ie, implicitly minimizes the trace of the covariance matrix for the conflicting task gradients, along with the Empirical Risk Minimization (ERM) gradient on the mean loss $\bar {\mathcal{L}}(\boldsymbol{\theta})$.
\end{theorem}

\begin{proof}
If ${\mathcal{L}_i}(\boldsymbol{\theta})$ has Lipschitz Hessian, \ie, $\|\nabla^{2}{\mathcal{L}_i}(\boldsymbol{\theta}_{1})-\nabla^{2}{\mathcal{L}_i}(\boldsymbol{\theta}_{2})\|\leq\rho\|\boldsymbol{\theta}_{1}-\boldsymbol{\theta}_{2}\|$ for some $\rho>0$, we can apply the fundamental theorem of Taylor's expansion to the gradient $\nabla{\mathcal{L}_i}(\boldsymbol{\theta}-\boldsymbol{v}_i(\boldsymbol{\theta}))$, then:
\begin{align*}
\nabla{\mathcal{L}_i}(\boldsymbol{\theta}-\boldsymbol{v}_i(\boldsymbol{\theta}))
=\nabla \mathcal{L}_i(\boldsymbol{\theta})-\nabla^2\mathcal{L}_i(\boldsymbol{\theta})\boldsymbol{v}_i(\boldsymbol{\theta})+\mathcal{O}(\|\boldsymbol{v}_i(\boldsymbol{\theta})\|^2).
\end{align*}
For instance, when $\boldsymbol{v}_i(\boldsymbol{\theta}) = \beta\left(\nabla \bar {\mathcal{L}}(\boldsymbol{\theta})-\nabla {\mathcal{L}_i}(\boldsymbol{\theta})\right)$, we have:
\begin{align*}
\nabla{\mathcal{L}_i}\left(\boldsymbol{\theta}-\beta(\nabla \bar {\mathcal{L}}(\boldsymbol{\theta})-\nabla {\mathcal{L}_i}(\boldsymbol{\theta}))\right)=\nabla \mathcal{L}_i(\boldsymbol{\theta})-\beta\nabla^2\mathcal{L}_i(\boldsymbol{\theta})(\nabla \bar {\mathcal{L}}(\boldsymbol{\theta})-\nabla {\mathcal{L}_i}(\boldsymbol{\theta}))+\mathcal{O}(\beta^2).
\end{align*}

Therefore, the update gradient $\boldsymbol{g}_{IGR}$ calculated after an displacement can be expressed as follows:
\begin{align*}
\boldsymbol{g}_{IGR}&=\frac{1}{m}\sum_{i=0}^{m-1}\nabla {\mathcal{L}_i}(\boldsymbol{\theta}-\boldsymbol{v}_i(\boldsymbol{\theta}))\\
&=\frac{1}{m}\sum_{i=0}^{m-1}\nabla \mathcal{L}_i(\boldsymbol{\theta})-\frac{\beta}{m}\sum_{i=0}^{m-1}\nabla^2\mathcal{L}_i(\boldsymbol{\theta})\left(\nabla \bar {\mathcal{L}}(\boldsymbol{\theta})-\nabla \mathcal{L}_i(\boldsymbol{\theta})\right)+\mathcal{O}(\beta^2)\\
&=\frac{1}{m}\sum_{i=0}^{m-1}\nabla \mathcal{L}_i(\boldsymbol{\theta})+\frac{\beta}m\sum_{i=0}^{m-1}(\nabla^2\mathcal{L}_i(\boldsymbol{\theta})\nabla \mathcal{L}_i(\boldsymbol{\theta})-\nabla^2\mathcal{L}_i(\boldsymbol{\theta})\nabla \bar {\mathcal{L}}(\boldsymbol{\theta}))+\mathcal{O}(\beta^2),
\end{align*}
given that $\sum_{i=0}^{m-1}(\nabla \mathcal{L}_i(\boldsymbol{\theta})-\nabla \bar {\mathcal{L}}(\boldsymbol{\theta}))=0$, then:
\begin{align*}
\boldsymbol{g}_{IGR}&=\nabla \bar {\mathcal{L}}(\boldsymbol{\theta})+\frac{\beta}m\sum_{i=0}^{m-1}(\nabla^2\mathcal{L}_i(\boldsymbol{\theta})\nabla \mathcal{L}_i(\boldsymbol{\theta})-\nabla^2\mathcal{L}_i(\boldsymbol{\theta})\nabla \bar {\mathcal{L}}(\boldsymbol{\theta}))+\mathcal{O}(\beta^2)\\
&=\nabla \bar {\mathcal{L}}(\boldsymbol{\theta})+\frac{\beta}{m}\sum_{i=0}^{m-1}(\nabla^2\mathcal{L}_i(\boldsymbol{\theta})\nabla \mathcal{L}_i(\boldsymbol{\theta})-\nabla^2\mathcal{L}_i(\boldsymbol{\theta})\nabla \bar {\mathcal{L}}(\boldsymbol{\theta})-\nabla^{2}\bar {\mathcal{L}}(\boldsymbol{\theta})\nabla \mathcal{L}_i(\boldsymbol{\theta})+\nabla^{2}\bar {\mathcal{L}}(\boldsymbol{\theta})\nabla \bar {\mathcal{L}}(\boldsymbol{\theta}))+\mathcal{O}(\beta^{2}) \\
&=\nabla \bar {\mathcal{L}}(\boldsymbol{\theta})+\frac{\beta}{m}\sum_{i=0}^{m-1}(\nabla^2\mathcal{L}_i(\boldsymbol{\theta})-\nabla^{2}\bar {\mathcal{L}}(\boldsymbol{\theta}))(\nabla \mathcal{L}_i(\boldsymbol{\theta})-\nabla\bar {\mathcal{L}}(\boldsymbol{\theta}))+\mathcal{O}(\beta^{2}) \\
&=\nabla \bar {\mathcal{L}}(\boldsymbol{\theta}) + \frac{\beta}{2m}\nabla(\sum_{i=0}^{m-1}\|\nabla \mathcal{L}_i(\boldsymbol{\theta})-\nabla \bar {\mathcal{L}}(\boldsymbol{\theta})\|^2)+\mathcal{O}(\beta^2).
\end{align*}
\end{proof}

\section{Empirical Observation Details}\label{app:rethink}
In the experiment (\cref{fig:accuracy gain}) on overlapping classes, we standardize the architecture of pre-trained models to Conv4.
Initially, we select a basic model pre-trained on 5 classes of \emph{CIFAR-FS}. Then, we choose pre-trained models with 0, 1, 2, 3, 4, and 5 overlapping training classes with the basic model, selecting 10 models for each category. In total, this amounts to 60 auxiliary models.
Tasks are recovered from pre-trained models based on \cref{eq:Loss_g} and used as training data for the meta-model. For the recovered task $\hat{\mathcal{T}}$ with class labels $\boldsymbol{Y}$, we randomly divide it into a support set $\boldsymbol{S}$ and a query set $\boldsymbol{Q}$. Then, we employ MAML~\cite{finn2017model} to optimize the meta-model $\boldsymbol{\theta}$:
\begin{equation}\label{eq:maml}
\begin{aligned}
\min_{\boldsymbol{\theta}}\mathbb{E}_{\hat{\mathcal{T}}}\mathcal{L}_{outer}
=&CE(M_{meta}(\boldsymbol{Q};\boldsymbol{\theta}_c),\boldsymbol{Y}_{Q}),\\
\mathrm{s.t.~}\boldsymbol{\theta}_{c}
=\min_{\boldsymbol{\theta}}\mathcal{L}_{inner}&=CE(M_{meta}(\boldsymbol{S};\boldsymbol{\theta}),\boldsymbol{Y}_{S}).
\end{aligned}
\end{equation}
Initially, we train the meta-model with data from the basic model $M_{bas}$ to establish a basic accuracy $\mathcal{P} (\boldsymbol{\theta}(M_\mathrm{bas}))$. The accuracy is calculated by standard few-shot classification on meta-testing subsets. To measure the impact of the auxiliary model $M_{aux}$ in joint training, we combine the losses of the basic and auxiliary models to update the meta-model, thereby obtaining a new accuracy $\mathcal{P} (\boldsymbol{\theta}(M_\mathrm{bas},M_\mathrm{aux}))$ and calculating the Accuracy Gain.

\begin{algorithm}[bht]
\small
\DontPrintSemicolon
\SetKwInOut{Input}{Input}\SetKwInOut{Output}{Output}\SetKwInOut{Require}{Require}
\textbf{Input: }{A collection of auxiliary models $\mathcal{M}_{pool}$; a basic model~$M_{\mathrm{bas}}$; a meta-model $M_{meta}(\cdot;\boldsymbol{\theta})$.
}\\
\textbf{Output: }{A list of the Accuracy Gain $[AG]$.}\\
Recover data from $M_{\mathrm{bas}}$ to train $M_{meta}(\cdot;\boldsymbol{\theta})$\\
Calculate $\mathcal{P} (\boldsymbol{\theta}(M_\mathrm{bas}) )$ on meta-testing subset\\
\For{ $M_{\mathrm{aux}}$ \rm\textbf{in} $\mathcal{M}_{pool}$}{
Re-initialize the meta-model $M_{meta}(\cdot;\boldsymbol{\theta})$\\
Recover training data from $M_{\mathrm{aux}}$\\
\For{ $\boldsymbol{\hat{X}}_{\mathrm{bas}}$,$\boldsymbol{\hat{X}}_{\mathrm{aux}}$ \rm \textbf{in} \rm {DataLoader}}{
Loss $= \mathrm{MAML}(\boldsymbol{\hat{X}}_{\mathrm{bas}};\boldsymbol{\theta}) + \mathrm{MAML}(\boldsymbol{\hat{X}}_{\mathrm{aux}};\boldsymbol{\theta})$\\
Loss.backward()\\
Update($M_{meta}(\cdot;\boldsymbol{\theta})$)
}
Calculate $\mathcal{P} (\boldsymbol{\theta}(M_\mathrm{bas},M_\mathrm{aux}))$ on meta-testing subset\\
$AG = \mathcal{P} (\boldsymbol{\theta}(M_\mathrm{bas},M_\mathrm{aux})) - \mathcal{P} (\boldsymbol{\theta}(M_\mathrm{bas}) )$ 
}
\caption{Pseudo-Code of the Empirical Observation}
\label{app:alg}
\end{algorithm}

In the experiment on model architecture, we keep the training classes of pre-trained models constant. The architecture of the basic model is Conv4, and we select pre-trained models with architectures of Conv4, ResNet-10, ResNet-18, and ResNet-50, 10 models for each category. The subsequent steps are the same as described above. We provide the pseudo-code for the empirical observation implementation in \cref{app:alg}.

\section{Task Grouping Strategies}\label{app:similar}
It is commonly believed that related tasks with similar underlying structures can benefit from training together~\cite{yu2020gradient,fifty2021efficiently}. However, we argue that leveraging the model heterogeneity by grouping dissimilar pre-trained models can also be beneficial. To validate whether our insight holds true, we conduct an ablation study comparing groupings based on both dissimilarity and similarity measures. Specifically, we derive the task similarity matrix $\left[ d_{\cos}\Big(\frac{\boldsymbol{F}^i}{\boldsymbol{F}^i+\boldsymbol{F}^j},\frac{\boldsymbol{F}^j}{\boldsymbol{F}^i+\boldsymbol{F}^j}\Big)\right]_{1 \leq i, j \leq n}$ to group the most similar pre-trained models together, keeping other steps of the process constant. As shown in \cref{table:similar}, groupings based on dissimilarity achieve better results. This further supports our insight that grouping and aligning dissimilar tasks can effectively balance the heterogeneity-homogeneity trade-off.

\begin{table}[bth]
\footnotesize
\centering
\renewcommand\arraystretch{1.1}
\setlength{\tabcolsep}{4.85pt}
\caption{Different task grouping strategies. Grouping pre-trained models based on dissimilarity or similarity measures.}
\resizebox{0.7\linewidth}{!}{
\begin{tabular}{lcccccc}
			\addlinespace
			\toprule
			\specialrule{0em}{1pt}{1pt}
			\multirow{2}{*}{ \bf Strategy} &\multicolumn{2}{c}{\emph {CIFAR-FS}} & \multicolumn{2}{c}{\emph {miniImageNet}} & \multicolumn{2}{c}{\emph {CUB}}\\ 
			\cmidrule(l){2-3} \cmidrule(l){4-5} \cmidrule(l){6-7}
			\specialrule{0em}{1pt}{1pt}
			&{$5$-way $1$-shot} & {$5$-way $5$-shot} &{$5$-way $1$-shot} &{$5$-way $5$-shot} &{$5$-way $1$-shot} &{$5$-way $5$-shot}\\
			\midrule
			\specialrule{0em}{1pt}{1pt}
			Similarity & {43.54} {\scriptsize $\pm$ 0.85} & {55.71} {\scriptsize $\pm$ 0.80} &{35.82} {\scriptsize $\pm$ 0.68}& {48.73} {\scriptsize $\pm$ 0.68}  &{33.50} {\scriptsize $\pm$ 0.62}& {46.77} {\scriptsize $\pm$ 0.67}\\
           \specialrule{0em}{1pt}{1pt}
           Dissimilarity   &   \bf {43.58 {\scriptsize $\pm$ 0.82}} & \bf {57.49 {\scriptsize  $\pm$ 0.79}}& \bf {36.20 {\scriptsize  $\pm$ 0.71}} & \bf {49.00 {\scriptsize  $\pm$ 0.71}}&   \bf{33.87 {\scriptsize $\pm$ 0.63}} & \bf {47.37 {\scriptsize  $\pm$ 0.73}}\\ 
           \bottomrule
	\end{tabular}}
\label{table:similar}
\end{table}


\end{document}